%% file: main.tex
\documentclass[acmsmall,nonacm,screen,11pt]{acmart}

\setcopyright{none}
\acmYear{2026}

\usepackage{brand}
\usepackage{header}
\usepackage{docspec}

\usepackage{amsfonts}
\usepackage{amssymb}
\usepackage{pifont}
\AtBeginDocument{}
\newcommand{\cmark}{\ding{51}}
\newcommand{\xmark}{\ding{55}}

\usepackage{nicefrac}
\usepackage{subcaption}
\usepackage{makecell}
\usepackage{colortbl}
\usepackage{wrapfig}
\usepackage{placeins}
\usepackage{algorithm}
\usepackage{algorithmic}
\usepackage{listings}
\usepackage{xspace}
\usepackage{tabularx}
\usepackage{multirow}
\usepackage{enumitem}
\usepackage[nameinlink,capitalise]{cleveref}

\graphicspath{{fig/}}

\newcommand{\method}{SHRED\xspace}

\begin{document}

\title[SHRED: Self-Play for Heterogeneous Road-Users without Expert Demonstrations]{\method: Self-Play for Heterogeneous\\Road-Users without Expert Demonstrations}

\author{Wei-Jer Chang}
\affiliation{\institution{Applied Intuition}\city{Mountain View}\state{CA}\country{USA}}
\author{Zhouchonghao Wu}
\affiliation{\institution{Applied Intuition}\city{Mountain View}\state{CA}\country{USA}}
\author{Akshay Rangesh}
\affiliation{\institution{Applied Intuition}\city{Mountain View}\state{CA}\country{USA}}
\author{Weixin Li}
\affiliation{\institution{Applied Intuition}\city{Mountain View}\state{CA}\country{USA}}
\author{Masayoshi Tomizuka}
\affiliation{\institution{University of California, Berkeley}\city{Berkeley}\state{CA}\country{USA}}
\author{Wei Zhan}
\affiliation{\institution{Applied Intuition}\city{Mountain View}\state{CA}\country{USA}}

\compactauthors{%
  {\raggedright
   {\rmfamily\bfseries
    Wei-Jer Chang\textsuperscript{1,*}, \;
    Zhouchonghao Wu\textsuperscript{1,*}, \;
    Akshay Rangesh\textsuperscript{1,*}, \;
    Weixin Li\textsuperscript{1},\\[0.25em]
    Masayoshi Tomizuka\textsuperscript{2}, \;
    Wei Zhan\textsuperscript{1,2}\par}
   \smallskip
   {\small\rmfamily\color{AppliedGray}%
    \textsuperscript{1}Applied Intuition \quad
    \textsuperscript{2}University of California, Berkeley\par}
   \smallskip
   {\footnotesize\rmfamily\color{AppliedGray}%
    \textsuperscript{*}Equal contribution.\par}
   \par}
}

\begin{abstract}

\input{sections/abstract}
\end{abstract}

\maketitle

\clearpage
\begingroup
\hypersetup{linkcolor=AppliedBlack}
\tableofcontents
\endgroup
\clearpage

\input{sections/intro}
\input{sections/related}

\input{sections/method}

\input{sections/experiments}

\input{sections/conclusion}

\FloatBarrier

\bibliographystyle{ACM-Reference-Format}
\bibliography{references}

\newpage
\appendix
\input{sections/appendix_method}

\end{document}

%% file: sections/abstract.tex
Realistic traffic simulation and autonomous planning both depend on modeling interactions among heterogeneous road users whose dynamics, constraints, and behaviors differ fundamentally. Existing approaches predominantly learn realistic interaction behavior from large-scale human driving demonstrations. In this work, we ask whether such behavior can instead emerge purely through self-play. We present \method, a framework that jointly trains vehicles, pedestrians, and cyclists through large-scale heterogeneous self-play from random initialization, without demonstrations or logged-trajectory replay. \method combines type-specific dynamics and objectives with procedural scenario generation and per-agent randomization, allowing diverse behaviors to emerge through closed-loop interaction.
On the Waymo Open Sim Agents Challenge (WOSAC), \method jointly controls all three road-user types and outperforms both the closest demonstration-free self-play baseline and prior self-play methods that use demonstration-trained reference policies. Policies trained only on nuPlan map geometry retain comparable realism under zero-shot evaluation on Waymo. For planning, SHRED outperforms all compared planners on the interaction-heavy InterPlan benchmark. Controlled ablations show that heterogeneity in self-play improves both traffic realism and planning performance, with particularly large planning gains in interaction-heavy scenarios. Together, these results establish heterogeneous procedural self-play as a scalable, demonstration-free paradigm for learning realistic and interactive traffic behavior.

%% file: sections/intro.tex
\section{Introduction}
\label{sec:intro}

Developing autonomous vehicles that operate safely alongside human road users requires scalable and realistic closed-loop traffic simulation. Real-world traffic, however, is inherently heterogeneous: beyond vehicles, it includes pedestrians, cyclists, and other agents with distinct physical dynamics, motion constraints, and behavioral patterns. These differences shape how agents respond to one another and how their interactions unfold over time. A pedestrian or scooter yielding at a crosswalk and a vehicle merging into traffic follow fundamentally different rules of engagement. Capturing such heterogeneous interactions is therefore central to realistic closed-loop simulation. It is equally central to autonomous planning, where the ego vehicle must anticipate how diverse agents will react to its own actions.

Recent advances in traffic simulation and autonomous planning have widely adopted data-driven learning, leveraging large-scale human driving datasets to model traffic behavior and train driving policies. Tokenized and generative simulation models reproduce realistic traffic behaviors and achieve strong performance on large-scale simulation benchmarks~\citep{wu2024smart,philion2024trajeglish,zhou2024behaviorgpt}. However, these approaches often rely on computationally expensive centralized models, limiting throughput in large-scale closed-loop simulation~\citep{chang2025spacer}. In parallel, learning-based planning methods have increasingly leveraged large-scale driving datasets to learn scene representations, model interactions, and optimize driving decisions, achieving strong performance across challenging driving scenarios~\citep{huang2023gameformer,huang2024dtpp,zheng2025diffplanner,tan2025flowplanner}. Despite their different objectives, both lines of work remain widely dependent on logged human driving data for learning realistic and effective interaction behaviors.
Their behavior remains tied to the interaction distributions in logged datasets, which can make coverage of diverse heterogeneous interactions and cross-dataset transfer more challenging. Large-scale demonstration datasets are also expensive to collect and may not fully capture the diversity of reactive multi-agent behaviors observed in real-world traffic.

Self-play reinforcement learning offers a complementary direction: it can learn through online interaction without trajectory supervision, scales naturally to large numbers of agents, and enables diverse multi-agent interactions to emerge through closed-loop rollouts.
Recent work has shown that self-play can produce robust driving policies at unprecedented scale~\citep{cusumano2025gigaflow} and achieve competitive realism when anchored to pretrained reference models~\citep{chang2025spacer}. However, prior self-play systems focus on vehicle policies or homogeneous dynamics and rewards across road-user types despite fundamentally different motion characteristics. Scalable, fully reactive heterogeneous self-play across vehicles, pedestrians, and cyclists without relying on logged trajectory data or pretrained imitation-policy anchors therefore remains largely unexplored.

To address these challenges, we propose \method~(\textbf{S}elf-Play for \textbf{H}eterogeneous \textbf{R}oad-Users without \textbf{E}xpert \textbf{D}emonstrations), a scalable framework for learning heterogeneous traffic behavior through large-scale self-play. \method jointly trains vehicles, pedestrians, and cyclists from random initialization, with type-specific dynamics, objectives, and action spaces that reflect their distinct modes of interaction. Procedural scenario generation and per-agent randomization of dynamics and reward parameters expose agents to diverse traffic configurations and behaviors throughout training. This enables a compact decentralized policy to learn realistic heterogeneous multi-agent behaviors entirely through self-play, without demonstration supervision.

\begin{figure}[t]
  \centering
  \includegraphics[width=\linewidth]{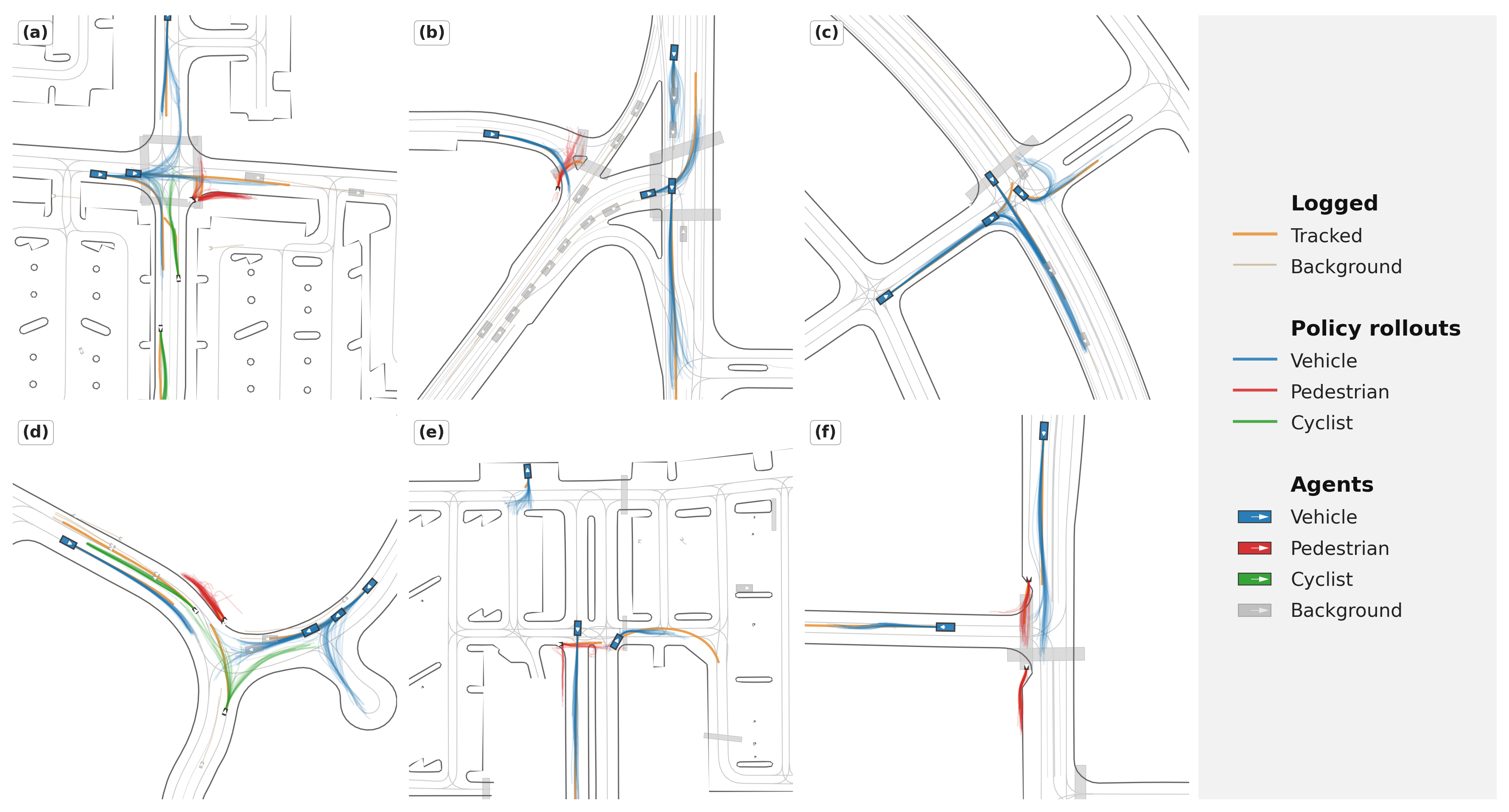}
  \caption{\textbf{Qualitative Rollout Distributions.} Qualitative trajectory distributions from \method across a diverse range of scenarios, from urban intersections to parking lots. Each panel shows 32 closed-loop policy rollouts (blue: vehicle, red: pedestrian, green: cyclist); orange traces show the ground-truth log trajectory. The panels show alternative maneuvers alongside the logged trajectory.}
  \label{fig:qualitative_dist}
\end{figure}

We show that demonstration-free self-play can achieve both realistic
heterogeneous traffic simulation and strong interactive planning. For
simulation, we evaluate on the Waymo Open Sim Agents Challenge
(WOSAC)~\citep{montali2023wosac} while jointly controlling vehicles, pedestrians,
and cyclists. \method outperforms the closest demonstration-free self-play
baseline, while controlled ablations show that heterogeneous self-play
substantially improves traffic realism, including the behavior of vehicles
themselves. For planning, we evaluate \method on
InterPlan~\citep{hallgarten2024interplan}, where it outperforms all compared planners
on interaction-heavy closed-loop scenarios. Complementary planning ablations
show that heterogeneous self-play substantially outperforms vehicle-only
self-play, with the largest gains on scenarios involving concentrated
interactions with other road users. Beyond these benchmarks, policies trained
on nuPlan maps transfer zero-shot to Waymo, while behavioral analyses show
that self-play produces diverse, plausible alternatives to logged trajectories
(Figure~\ref{fig:qualitative_dist}).

Our main contributions are as follows:
\begin{itemize}
    \item We introduce \method, a demonstration-free heterogeneous self-play framework that jointly learns vehicle, pedestrian, and cyclist behaviors with distinct dynamics and objectives, entirely without demonstrations or logged-trajectory replay.

    \item We develop the key components for scalable heterogeneous self-play, including procedural scenario and goal generation, per-agent randomization of dynamics and objectives, and type-specific policy components that support joint learning across diverse road-user types.

    \item We show that demonstration-free self-play is effective across both simulation and planning: \method achieves strong sim-agent realism on WOSAC, the highest score among compared planners on the official 80-scenario InterPlan split, and zero-shot map transfer from nuPlan to Waymo.
\end{itemize}

%% file: sections/related.tex
\section{Related Work}
\label{sec:related}
\subsection{Data-Driven and Heterogeneous Traffic Simulation}
Imitation from human driving logs has become the dominant paradigm for
realistic traffic simulation.
Tokenized autoregressive
models~\citep{wu2024smart, philion2024trajeglish,
zhou2024behaviorgpt, zhang2025catk} and diffusion-based
approaches~\citep{jiang2024scenediffuser} have driven strong performance
on WOSAC~\citep{montali2023wosac}, while closed-loop fine-tuning,
promptable rollout, and RL/RLHF post-training further improve rollout
realism and controllability~\citep{zhang2025catk, tan2024prosim,
peng2024improving, chen2025rift, cao2024rlhf, pei2025smartr1}.
RLFTSim~\citep{ahmadi2026rlftsim} fine-tunes a pretrained traffic simulator with a distribution-alignment reward. DecompGAIL~\citep{guo2026decompgail} combines a SMART-based model with decomposed adversarial imitation learning. Both methods use a trajectory-model prior; SHRED instead studies heterogeneous recurrent self-play without one.
Pedestrian and cyclist behavior has additionally been studied through
rule-based crowd dynamics~\citep{helbing1995social,
helbing2005crowd, vandenberg2008orca}, multimodal trajectory forecasting
models~\citep{alahi2016sociallstm, gupta2018socialgan,
salzmann2020trajectronpp, xu2022socialvae,
rasouli2023pedformer}, and multi-agent driving
simulators~\citep{suo2021trafficsim, vinitsky2022nocturne,
gulino2023waymax, li2022metadrive, kazemkhani2024gpudrive,
rowe2024ctrlsim}.
However, realistic heterogeneous interaction behavior is typically
learned implicitly from demonstrations, while learned closed-loop
control remains largely vehicle-centric or relies on shared dynamics
across all road users.
Our work instead studies whether realistic heterogeneous traffic can
emerge from demonstration-free self-play with type-specific dynamics,
rewards, goals, and policy components for vehicles, pedestrians, and
cyclists jointly.

\subsection{Self-Play Reinforcement Learning for Driving}
Self-play has driven landmark results in game
playing~\citep{silver2017mastering, vinyals2019grandmaster,
berner2019dota, jaderberg2019human} and has recently been adopted for
traffic simulation.
Gigaflow~\citep{cusumano2025gigaflow} is the closest demonstration-free
precedent, training a unified policy for vehicles, pedestrians, and
cyclists through massively parallel self-play.
However, all agents share the same jerk-actuated bicycle dynamics and
goal-generation mechanism, with diversity induced primarily through
conditioning variables inside a shared backbone.
Other self-play systems explore complementary directions:
\citet{cornelisse2025reliable} scale vehicle-only PPO
from logged WOMD initial states, while HR-PPO~\citep{cornelisse2024hrppo}
and SPACeR~\citep{chang2025spacer} regularize self-play with
behavior-cloned or tokenized policy anchors.
NOMAD~\citep{wang2026nomad} studies map-transfer through self-play, and
GPUDrive~\citep{kazemkhani2024gpudrive} provides a GPU-resident substrate
for large-scale simulation.
TerraTransfer~\citep{xiong2026terratransfer} aligns a self-play policy with a
pretrained vision backbone to obtain end-to-end driving policies without
expert demonstrations.
In contrast, we jointly train vehicles, pedestrians, and cyclists from
fully procedural initialization, without trajectory supervision, pretrained imitation-policy anchors, or logged-trajectory replay, using type-specific dynamics,
rewards, and goal generation.

\subsection{Heterogeneity in Multi-Agent Reinforcement Learning}
Most heterogeneous multi-agent RL methods are developed for cooperative
settings with shared rewards, including QMIX~\citep{rashid2018qmix},
MAPPO~\citep{yu2022mappo}, and HAPPO/HATRPO~\citep{kuba2022happo,
zhong2024harl}. Several of them rely on sequential per-agent updates
(HAPPO/HATRPO) or separate per-agent policies, which become costly in
traffic scenes containing dozens of agents with different dynamics and
action spaces.
Recent modular and role-conditioned approaches such as
SePS~\citep{christianos2021seps}, ROMA~\citep{wang2020roma},
RODE~\citep{wang2021rode}, and MoE-based RL
architectures~\citep{yang2020softmod,obandoceron2024moe} reduce
interference through selective parameter sharing or latent routing, but
remain primarily focused on cooperative domains.

Our setting is instead mixed-motive: each road user optimizes its own
goal under type-specific rewards and dynamics without a shared team
objective.
We therefore organize policies by physical agent type rather than agent
identity, sharing perception while maintaining separate recurrent,
actor, and value heads for vehicles, pedestrians, and cyclists.

%% file: sections/method.tex
\section{Method}
\label{sec:method}
\subsection{Heterogeneous Agent Modeling}
\label{sec:method:agent}

We consider heterogeneous traffic environments containing agent types
$\mathcal{K}=\{\mathrm{VEH},\mathrm{PED},\mathrm{CYC}\}$,
corresponding to vehicles, pedestrians, and cyclists.
Each agent type follows its own dynamics model, reward structure,
and discrete action space while interacting through a shared closed-loop
environment.
At each timestep, agent $i$ selects an action according to a decentralized
policy
\[
a_t^i \sim \pi_\theta\!\left(
o_t^i,\,
g_t^i,\,
k^i,\,
C_{\mathrm{dyn}}^i,\,
\alpha^i
\right),
\]
where $o_t^i$ denotes the local observation,
$g_t^i$ denotes goal and waypoint conditioning,
$k^i$ is the agent type,
$C_{\mathrm{dyn}}^i$ denotes kinematic parameters,
and $\alpha^i$ parameterizes the reward function.

\paragraph{Per-Type Dynamics and Actions.}
Each agent type follows its own continuous-control dynamics model.
Vehicles use jerk-integrated Ackermann dynamics, pedestrians follow
unicycle dynamics, and cyclists use kinematic bicycle dynamics.
Actions are discretized separately for each type to reflect their distinct
motion characteristics and control regimes.
Vehicles operate over longitudinal and lateral jerk primitives,
pedestrians use velocity and turning controls with an explicit stop action,
and cyclists follow acceleration and steering controls. Appendix~\ref{app:dynamics} gives the full dynamics and action grids.

\paragraph{Rewards.}
\emph{Vehicle} rewards inherit the Gigaflow~\citep{cusumano2025gigaflow} formulation, while pedestrian and cyclist rewards introduce type-specific behavioral priors for heterogeneous traffic interaction.
\emph{Pedestrian} rewards remove lane-following objectives and instead encourage sidewalk, crosswalk, and waiting behavior through soft road-intrusion penalties and walking-speed regularization, allowing curb waiting, yielding, and occasional jaywalking while discouraging unrealistic motion.
\emph{Cyclist} rewards retain vehicle-like navigation objectives while introducing soft lane-edge preferences and relaxed lane-alignment constraints, encouraging curb-side riding without enforcing strict vehicle lane-centering behavior.

We also introduce asymmetric cross-type collision shaping: a vehicle's
collision penalty doubles when the other party is a pedestrian. Detailed reward formulations, thresholds, and parameter ranges are
provided in Appendix~\ref{app:rewards}.

\begin{figure}[!htbp]
\centering
\includegraphics[width=\textwidth]{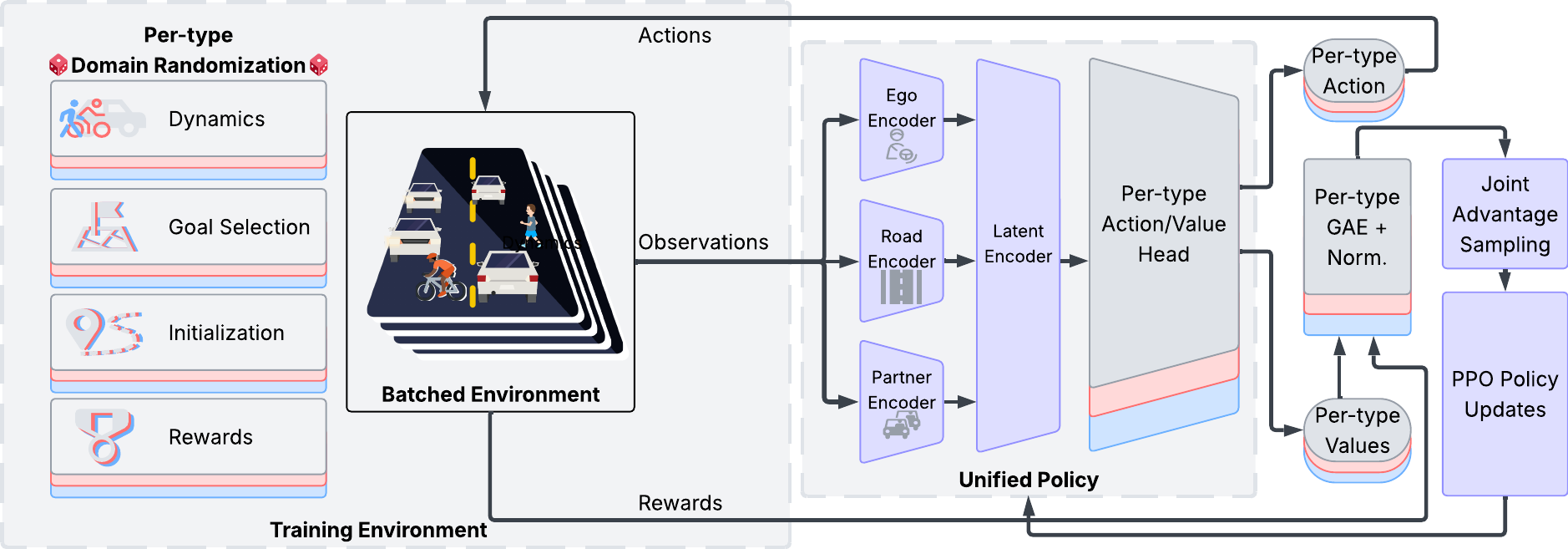}
\caption{\textbf{\method Training Architecture.} Simulation produces mixed transitions across all agent types. The model consists of shared encoders followed by per-type LSTM cells, action heads, and value heads. Per-type rewards and value normalization accommodate distinct reward scales across road-user types. Each type uses its own dynamics model: jerk-integrated Ackermann dynamics for vehicles, unicycle dynamics for pedestrians, and kinematic bicycle dynamics for cyclists.}
\label{fig:training_arch}
\end{figure}

\subsection{Procedural Heterogeneous Self-Play}
\label{sec:method:selfplay}

Recent self-play works often initialize agents directly
from logged datasets or replay recorded trajectories during training
and evaluation~\citep{cornelisse2025reliable,chang2025spacer}.
While effective for reproducing realistic traffic scenes, such approaches
remain constrained by the finite scene layouts and trajectory horizons
present in recorded data.
To improve coverage and interaction diversity during heterogeneous
self-play, we procedurally generate closed-loop training episodes through
randomized initialization, goal generation, and agent parameterization.

Formally, each training episode is sampled from a factorized procedural
distribution
\[
\mathcal{D}_{\mathrm{ep}}
=
\mathcal{D}_{\mathrm{scene}}
\prod_{k\in\mathcal{K}}
\prod_{i\in\mathcal{N}_k}
\mathcal{D}^{(k)}_{\mathrm{init}}
\mathcal{D}^{(k)}_{\mathrm{goal}}
\mathcal{D}^{(k)}_{b}
\mathcal{D}^{(k)}_{C_{\mathrm{dyn}}}
\mathcal{D}^{(k)}_{\alpha}
\mathcal{D}_{\mathrm{mask}},
\]
where each factor is parameterized independently per agent type.
These factors jointly control the distribution of scene layouts,
navigation objectives, agent behaviors, and goal observability
encountered during self-play training.

\paragraph{Per-Type Initialization ($\mathcal{D}_{\mathrm{init}}^{(k)}$).}
Controlled agents are initialized procedurally from type-specific spawn
surfaces.
Vehicles and cyclists are sampled from lane centerlines, while pedestrians
are initialized from crosswalks and road-edge regions.
Agents are added sequentially through rejection sampling to ensure
collision-free initial states, with randomized heading perturbations
applied at spawn time to increase local interaction diversity under
high environment densities.
Because initialization is performed independently for each agent type,
the resulting scenes span diverse heterogeneous compositions.
These include dense crosswalk traffic, mixed cyclist--vehicle flows, and
multi-type urban intersections, which are rare in recorded datasets.

\paragraph{Procedural Goal Generation.}
Rather than replaying logged destinations, goals are procedurally sampled
from map topology conditioned on agent type.
Vehicles and cyclists follow lane-graph connectivity, while pedestrian
goals are generated along crosswalk and sidewalk topology.
Unlike logged-goal initialization, where agents often terminate after
reaching a single recorded destination, goals are continuously regenerated
after completion during training.
This decouples interaction horizon from the finite duration of recorded
scenarios, allowing agents to experience substantially longer and more
diverse closed-loop interactions within a single environment rollout.

\paragraph{Per-Agent Diversity.}
Behavioral diversity is induced through independent randomization of
agent geometry, kinematic coefficients, and reward parameterizations.
Each agent observes its own
$(b^i, C_{\mathrm{dyn}}^i, \alpha^i)$
but not those of surrounding agents, forcing the policy to model partner
behavior as arising from a heterogeneous population rather than a fixed
shared dynamics model.

\paragraph{Goal Dropout.}
To reduce over-reliance on explicit navigation conditioning, a subset of
agents have their goal observations masked during training.
These agents must instead act using local scene context and interaction
history alone, while continuing to participate in fully reactive
closed-loop self-play.
This encourages the policy to produce robust and realistic behaviors even
when long-horizon navigation intent is partially specified or absent.

\subsection{Architecture and Algorithm Design Choices}
\label{sec:method:selfplay:optimization}

\paragraph{Policy Architecture.}
The architecture of \method is illustrated in
Figure~\ref{fig:training_arch}.
All agent types share encoders for ego state, road geometry, nearby agents, traffic signals, goals, and conditioning variables. The shared encoders produce a scene representation for each agent. Vehicles, pedestrians, and cyclists then use separate LSTM cells, actor heads, and value heads.

\paragraph{Value Normalization and Advantage Computation.}
Jointly optimizing heterogeneous agent populations introduces substantial
optimization imbalance: different agent types exhibit distinct reward
scales, motion statistics, and advantage distributions during self-play.
As a result, naively optimizing all agents under a shared PPO objective
can cause dominant agent classes or low-signal transitions to overwhelm
the optimization process.

To stabilize heterogeneous self-play, we maintain independent PopArt
normalization statistics~\citep{vanhasselt2016popart} together with
per-type advantage standardization before PPO updates.
Advantages are computed jointly over mixed heterogeneous rollouts using
GAE with V-trace corrections~\citep{espeholt2018impala}.
We further perform advantage-weighted prioritized replay,
replacing hard advantage filtering used in prior large-scale self-play
systems~\citep{cusumano2025gigaflow}, where rollout segments are sampled
proportionally to the magnitude of their estimated advantages.
This increases the contribution of high-signal transitions during
optimization while preserving balanced learning across heterogeneous
agent populations.

%% file: sections/experiments.tex
\section{Experiments}
\label{sec:experiments}

We evaluate demonstration-free heterogeneous self-play across both simulation and planning, focusing on its role in learning realistic traffic behavior and effective interactive planning.

\subsection{Heterogeneous Traffic Simulation}
\label{sec:exp:sim_agent}

\paragraph{Training and Evaluation Protocol.}
We train the recurrent heterogeneous policy in the TerraZero simulator~\citep{wu2026terrazero} with distributed mixed-precision self-play on 32 NVIDIA A100 GPUs. The policy shares scene encoders across road users and uses separate LSTM, actor, and value components for vehicles, pedestrians, and cyclists. Training uses map geometry, procedural initialization, and map-conditioned goals without trajectory supervision or a pretrained imitation-policy anchor. We evaluate WOSAC 2023 and WOSAC 2024 validation with 32 stochastic closed-loop rollouts per scenario. The WOSAC 2023 evaluation follows the leaderboard setting: agents receive the logged history and the map but no goal. Appendix~\ref{app:wosac} describes the evaluator verification and reports the WOSAC 2025 test evaluation separately.

\paragraph{Demonstration-Free Self-Play Comparison.}
Table~\ref{tab:wosac2023_demofree} compares \method with Gigaflow, the closest demonstration-free self-play baseline, on WOSAC 2023. Gigaflow applies its learned policy to vehicles and cyclists during evaluation while controlling pedestrians with a scripted IDM-like policy~\citep{cusumano2025gigaflow}. In contrast, \method learns policies for all three road-user types through self-play and achieves higher realism than Gigaflow (0.631 vs.\ 0.619) over 10,000 validation scenarios, with a particularly strong improvement on an interaction metric, time-to-collision likelihood (0.894 vs.\ 0.808). Gigaflow achieves a higher distance-to-object likelihood.

To examine whether the gain arises from heterogeneous self-play rather than self-play alone, we additionally evaluate a homogeneous \method variant with shared dynamics, rewards, and policy components across road-user types, which is the joint homogeneous design of Table~\ref{tab:ablation}. The homogeneous variant scores 0.604, compared with 0.631 for the full
heterogeneous policy. The largest degradations appear in motion realism,
with drops in linear-speed and angular-acceleration likelihood. Vehicles exhibit the largest per-type realism drop (0.641 to 0.614), suggesting that the benefits of heterogeneous modeling extend across road-user types.
\input{tab/tab2_gigaflow}

\paragraph{Role of Heterogeneity.}
\label{sec:exp:analysis}
\input{tab/tab_ablation}
Table~\ref{tab:ablation} examines how the components of \method's heterogeneous
modeling contribute to simulation realism. Homogeneous dynamics give every road-user type the pedestrian unicycle model, and homogeneous rewards give every type the vehicle reward set. Homogenizing either dynamics or rewards causes modest degradation, while the fully homogeneous design produces the largest drop in realism. Vehicles benefit the most from heterogeneous modeling. Removing procedural initialization and goals lowers overall realism by 0.010, with the largest drop for vehicles.

\input{tab/tab1}

\paragraph{Self-Play Realism, Diversity, and Map Transfer.}
Table~\ref{tab:wosac_main} compares \method with prior self-play approaches
on WOSAC 2024 across realism and trajectory diversity. We measure diversity as the mean pairwise Wasserstein distance between each agent's occupancy distributions across 32 rollouts, following the trajectory-space analysis of BITS~\citep{xu2023bits}. Notably, unlike HR-PPO
and SPACeR, \method achieves competitive realism and greater trajectory
diversity without demonstration data or a pretrained imitation-policy anchor.
Moreover, the nuPlan-trained policy retains comparable realism under zero-shot
evaluation on Waymo, suggesting that the learned traffic behaviors transfer
across maps and datasets.

\subsection{Interactive Ego Planning}
\label{sec:exp:interplan}

In this section, we evaluate whether policies learned through demonstration-free
heterogeneous self-play can support interactive ego planning. We use
InterPlan~\citep{hallgarten2024interplan} because it emphasizes challenging
interactions with other road users, particularly vulnerable road users (VRUs),
through long-tail scenarios including jaywalking pedestrians, overtaking,
dense lane changes, construction zones, and accident sites. The ego planner
uses the \method policy trained on nuPlan maps. To isolate the effect of
heterogeneous self-play, we compare against a separately trained vehicle-only
self-play policy in the planning ablation.

\input{tab/tab_interplan}

\paragraph{Planning Performance.}
Table~\ref{tab:interplan} presents representative baselines;
Appendix~\ref{app:interplan_full} reports the full comparison.
Beyond learning realistic heterogeneous traffic behavior, \method also
learns a robust ego planner, achieving the highest score among the compared
methods on both InterPlan evaluation sets: 67.50 on the official 80-scenario
split and 70.14 on the full 335-scenario evaluation. Notably, \method internalizes driving behavior through large-scale heterogeneous self-play in a compact 6.7M-parameter policy, orders of magnitude smaller than learned baselines such as the 7B-parameter LLaMA backbone used by HybridLLMPlanner, while achieving substantially higher performance (67.50 vs.\ 53.00). Even this 6.7M parameter count reflects a deliberate latency--memory tradeoff: fused heterogeneous LSTM weights increase parameter count and memory usage to enable lower-latency inference.

\paragraph{Planning Ablations.}
Table~\ref{tab:interplan} asks whether heterogeneous self-play benefits the
ego planner itself. We compare \method with a vehicle-only self-play policy
trained on the same nuPlan maps with the same vehicle dynamics and rewards.
The vehicle-only policy learns vehicle control through self-play while
pedestrians follow scripted crossing and jaywalking behaviors and cyclists
are absent, mirroring the general setup of prior vehicle-centric self-play
approaches such as Gigaflow~\citep{cusumano2025gigaflow}. \method improves the InterPlan score from
54.19 to 67.50 on the official 80-scenario split and from 66.65 to 70.14 on
the full 335-scenario evaluation. The substantially larger gain on the
official split, which concentrates challenging interactions with other road
users, suggests that heterogeneous self-play is particularly beneficial for
interactive planning.

\subsection{Behavioral Analysis}
\label{sec:exp:behavior}

\paragraph{Qualitative Evaluation of Data Diversity.}
\label{sec:exp:diversity}

A key question for any simulation policy is whether it reproduces only the
specific behaviors present in the training data, or whether it generalizes to a
broader distribution of plausible maneuvers.
To assess this, we generate 32 closed-loop rollouts per scenario and overlay
the resulting trajectory distributions against the ground-truth log
(Figure~\ref{fig:qualitative_dist}).
The policy distribution covers the log trajectory as a high-probability
mode while placing meaningful probability mass on alternative maneuvers:
unexecuted turns at intersections, alternative branches at forks,
and a range of merge timings. Supplementary videos show planning around vulnerable road users and in dense interactive scenarios.

\begin{figure}[!htbp]
\centering
\begin{minipage}[c]{0.58\linewidth}
\centering
\includegraphics[width=\linewidth]{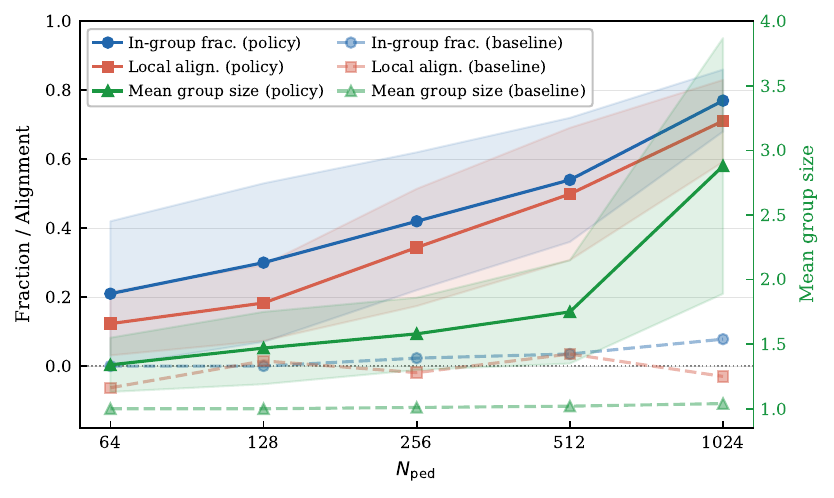}
\end{minipage}\hfill
\begin{minipage}[c]{0.41\linewidth}
\captionsetup{justification=raggedright,singlelinecheck=false}

\caption{\textbf{Group Behavior \mbox{Metrics}.} Proximity clustering and directional alignment versus pedestrian population $N_{\rm ped}$. Shaded bands: $\pm 1$ standard deviation across scenarios. Left axis: \textcolor[HTML]{2166ac}{in-group fraction $f_{\rm grp}$}; \textcolor[HTML]{d6604d}{local alignment $\mu_{\rm align}$}; right axis: \textcolor[HTML]{1a9641}{mean group size $\bar{g}$}. All three GBMs increase monotonically with crowd density for the policy while remaining near zero\,/\,unity for the stationary baseline.}
\label{fig:crowd_gbm_waymo}
\end{minipage}
\end{figure}

\paragraph{Emergent Collective Behavior.}
\label{sec:exp:crowd}

We observe that structured collective pedestrian behavior emerges from purely decentralized, self-interested agents without any group reward or shared objective.
To test this, we sweep the configured pedestrian
population $N_{\rm ped}\in\{64,\,128,\,256,\,512,\,1024\}$ while holding vehicles~($V=16$) and cyclists~($C=8$) fixed and evaluating over 32 validation scenarios of 60\,s.
We make no changes to the individual reward or goal structure; pedestrians independently optimize their own progress and safety objectives under random goal assignments.

We measure three \emph{Group Behavior Metrics} (GBMs):
{i)}~\emph{in-group fraction} $f_{\rm grp}$~\citep{kupper2023social}, the fraction of
pedestrians that spend at least 50\% of a rollout within 5\,m of at least one
co-moving companion (proximity-linked group);
{ii)}~\emph{local alignment} $\mu_{\rm align}$~\citep{luquesanchez2020crowd}, per-frame mean
cosine similarity between each pedestrian's velocity and those of its
$k$-nearest neighbors, averaged over moving frames; and
{iii)}~\emph{mean group size} $\bar{g}$~\citep{kupper2023social}, average headcount of
identified proximity groups.
A \emph{stationary baseline}, pedestrians with random positions and
orientations (used as proxy velocities) frozen for the entire rollout, provides a nonmoving reference.

Figure~\ref{fig:crowd_gbm_waymo} shows increasing proximity clustering and directional alignment as pedestrian density rises, despite the absence of an explicit group objective.
The stationary reference stays near its floor at every density, so static crowding alone does not produce the measured structure.
The result illustrates aggregate structure under individual objectives, consistent with the emergence of crowd patterns in local-interaction models~\citep{helbing1995social,moussaid2011simple}.

%% file: tab/tab2_gigaflow.tex
\begin{table}[t]
\centering
\caption{\textbf{Comparison with demonstration-free self-play on WOSAC 2023.} SHRED outperforms Gigaflow in overall realism, with the largest gains in time-to-collision and linear speed. Without type-specific rewards and dynamics, SHRED falls below Gigaflow, showing that its advantage comes from heterogeneous modeling. Neither method observes logged destinations and goals are sampled from the map, following the official WOSAC setting. Gigaflow controls pedestrians with a scripted IDM-like policy, whereas SHRED learns all three road-user types.}

\label{tab:wosac2023_demofree}
\footnotesize
\setlength{\tabcolsep}{1.2pt}
\resizebox{\linewidth}{!}{%
\begin{tabular}{@{}lcccccccccc@{}}
\toprule
Method
 & \makecell{Realism\\$\uparrow$}
 & \makecell{Lin.\\Speed $\uparrow$}
 & \makecell{Lin.\\Accel.\ $\uparrow$}
 & \makecell{Ang.\\Speed $\uparrow$}
 & \makecell{Ang.\\Accel.\ $\uparrow$}
 & \makecell{Dist.\\Obj.\ $\uparrow$}
 & \makecell{TTC\\$\uparrow$}
 & \makecell{Dist.\\Rd.\ Edge $\uparrow$}
 & \makecell{Off-\\road $\uparrow$}
 & \makecell{Coll.\\$\uparrow$} \\
\midrule
Gigaflow~\citep{cusumano2025gigaflow}
 & 0.619 & 0.261 & \textbf{0.253} & \textbf{0.506} & \textbf{0.476} & \textbf{0.316} & 0.808 & 0.545 & \textbf{0.910} & \textbf{0.947} \\
\midrule
\method
 & \textbf{0.631} & \textbf{0.353} & 0.247 & 0.501 & 0.457 & 0.250 & \textbf{0.894} & \textbf{0.607} & 0.904 & \textbf{0.947} \\
\makecell[l]{\method (homogeneous\\rewards and dynamics)}
 & 0.604 & 0.247 & 0.239 & 0.495 & 0.358 & 0.248 & 0.879 & 0.563 & 0.896 & \textbf{0.947} \\
\bottomrule
\end{tabular}%
}
\end{table}

%% file: tab/tab_ablation.tex
\begin{table}[t]
\centering
\caption{\textbf{Ablation of heterogeneous dynamics and rewards on WOSAC 2024.} Overall and per-type realism for vehicles, pedestrians, and cyclists. Removing either component modestly reduces realism, while removing both causes the largest drop, most strongly for vehicles. Removing random initialization and goals also degrades realism.}
\label{tab:ablation}
\small
\begin{tabular}{cccccc}
\toprule
Het.\ dyn. & Het.\ rew. & Overall $\uparrow$ & Veh. $\uparrow$ & Ped. $\uparrow$ & Cyc. $\uparrow$ \\
\midrule
\cmark & \cmark & \textbf{0.721} & \textbf{0.742} & \textbf{0.724} & 0.672 \\
\xmark & \cmark & 0.718 & 0.739 & 0.723 & 0.665 \\
\cmark & \xmark & 0.714 & 0.733 & 0.711 & 0.663 \\
\xmark & \xmark & 0.686 & 0.699 & 0.708 & 0.668 \\
\midrule
\multicolumn{2}{c}{w/o rand.\ init.\ \& goals} & 0.711 & 0.725 & 0.718 & \textbf{0.676} \\
\bottomrule
\end{tabular}
\end{table}

%% file: tab/tab1.tex
\begin{table}[t]
\centering
\caption{\textbf{Realism and diversity on WOSAC 2024.} Unlike HR-PPO and SPACeR, which anchor self-play to demonstration-trained policies, SHRED uses no demonstrations, yet achieves the highest interactive score and higher realism than both, with greater rollout diversity. The nuPlan-trained policy retains comparable realism zero-shot on Waymo.}

\label{tab:wosac_main}
\footnotesize
\setlength{\tabcolsep}{3pt}
\begin{tabular}{@{}lccccccc@{}}
\toprule
& & \multicolumn{4}{c}{WOSAC 2024} & \multicolumn{1}{c}{Diversity} \\
\cmidrule(lr){3-6}\cmidrule(l){7-7}
Method & \makecell{Demo\\Free} & Realism $\uparrow$ & Kinematic $\uparrow$ & Interactive $\uparrow$ & Map $\uparrow$ & \makecell{Mean dist. (m)\\$\uparrow$} \\
\midrule
\rowcolor[gray]{0.93}\makecell[l]{SMART (CAT-K)\\[-1pt]\citep{wu2024smart,zhang2025catk}}
& \xmark & 0.752 & 0.453 & 0.715 & 0.815 & 6.97 \\
\midrule
\makecell[l]{HR-PPO~\citep{cornelisse2024hrppo}}
& \xmark & 0.690 & 0.308 & 0.731 & 0.855 & 4.19 \\
\makecell[l]{SPACeR~\citep{chang2025spacer}}
& \xmark & 0.717 & 0.378 & 0.754 & \textbf{0.863} & 1.41 \\
\midrule
\method
& \cmark & \textbf{0.721} & \textbf{0.402} & \textbf{0.778} & 0.832 & \textbf{5.38} \\
\method (zero-shot, nuPlan)
& \cmark & 0.715 & 0.395 & 0.776 & 0.821 & 4.93 \\
\bottomrule
\end{tabular}
\end{table}

%% file: tab/tab_interplan.tex
\begin{table}[t]
\centering
\caption{\textbf{Interactive planning on InterPlan.} \method achieves the highest score among the compared planners on both the official 80-scenario split and the full 335-scenario evaluation. Comparison with a vehicle-only variant, trained with scripted pedestrians and no cyclists, showcases the benefit of heterogeneous self-play for interactive planning. Table~\ref{tab:interplan_full} in Appendix~\ref{app:interplan_full} gives the full comparison.}
\label{tab:interplan}
\scriptsize
\setlength{\tabcolsep}{2pt}
\resizebox{\linewidth}{!}{%
\begin{tabular}{@{}llcc@{}}
\toprule
Planner & Type & InterPlan (80) $\uparrow$ & Full Scale (335) $\uparrow$ \\
\midrule
IDM~\citep{treiber2000idm} & Rule & 31 & --- \\
SPDM ($N_p{=}60$)~\citep{distelzweig2026spdm} & Rule & 63.66 & --- \\
HybridLLMPlanner (Llama-7B)~\citep{hallgarten2024interplan} & Hybr. & 53 & --- \\
PDM+PPO~\citep{behaviorbenchmark2026} & Hybr. & 43.2 & --- \\
Diffusion Planner~\citep{zheng2025diffplanner} & IL & 25.8 & --- \\
FlowDrive~\citep{wang2025flowdrive} & IL & 36.96 & --- \\
Flow Planner~\citep{tan2025flowplanner} & IL & --- & 61.82 \\
PPO~\citep{behaviorbenchmark2026} & RL & 42.1 & --- \\
\midrule
\method & RL & \textbf{67.50} & \textbf{70.14} \\
\method (vehicle-only) & RL & 54.19 & 66.65 \\
\bottomrule
\end{tabular}%
}
\end{table}

%% file: sections/conclusion.tex
\section{Conclusion}
\label{sec:conclusion}

We have presented \method, a framework for jointly training vehicles,
pedestrians, and cyclists through self-play reinforcement learning with
type-specific dynamics, rewards, and value normalization, from random
initialization without trajectory supervision or logged-trajectory replay.
On the Waymo Open Sim Agents Challenge, \method achieves strong realism
without trajectory-supervised policy training, while policies trained on
nuPlan maps retain comparable realism under zero-shot evaluation on Waymo.
Beyond simulation, the learned policy also serves as an effective ego planner,
outperforming all compared planners on InterPlan.
Controlled ablations further show that heterogeneous self-play improves both
traffic realism and planning performance, with particularly large planning
gains in interaction-heavy scenarios.

Our results position heterogeneous self-play as a complementary paradigm to
data-driven approaches across both traffic simulation and autonomous planning.
Imitation-based methods retain an advantage in overall distributional realism
on WOSAC, while self-play removes dependence on demonstration data and enables behaviors
to emerge through closed-loop interaction. Together, our results suggest that heterogeneous self-play provides a scalable, demonstration-free paradigm for learning interactive traffic behavior across both simulation and planning.

\paragraph{Limitations and Future Work.}
SHRED inherits the fidelity ceiling of its simulator and reward design:
behaviors not captured by the dynamics or objectives, such as richer
right-of-way conventions and traffic rules, may not emerge through self-play.
Future work could address these limitations through richer traffic-rule
objectives, automatically learned or calibrated rewards, and additional
road-user types. A realism gap to imitation-based models also remains,
particularly in map-based realism (0.832 vs.\ 0.863 for SPACeR), suggesting
that lightweight data-driven priors could complement self-play where
demonstrations are available. Finally, WOSAC measures agreement with logged
behavior rather than behavioral validity, and we do not evaluate transfer to
real-world driving. Evaluating learned behaviors against real-world
interactions and using SHRED as interactive background traffic for downstream
planner training and evaluation are important directions for future work.

%% file: sections/appendix_method.tex
\section{Per-Type Dynamics}
\label{app:dynamics}

This section gives the full continuous-time integrators for each of the
three dynamics models summarized in \S\ref{sec:method:agent}.
All models share a common kinematic state
$s_t=(x,y,\theta,v_x,v_y,a_{\text{long}},a_{\text{lat}},\delta)$
and operate at $\Delta t = 0.1\,$s.
Per-type kinematic parameters and domain-randomization ranges appear in
Table~\ref{tab:kinematic_params}.

\subsection{Vehicle: Jerk-Integrated Ackermann Arc}

The vehicle model operates at the \emph{jerk} level: the policy selects
longitudinal and lateral jerk values from a discrete grid
$\dot{a}_{\text{long}} \in \{-15,\,-4,\,0,\,4\}$ m/s$^3$ and
$\dot{a}_{\text{lat}} \in \{-4,\,0,\,4\}$ m/s$^3$, giving $|\mathcal{A}|=12$
actions.

\paragraph{Acceleration Integration.}
Per-episode coefficients $C_{\text{throttle}},C_{\text{steer}}$ scale the
jerk inputs:
\begin{align}
a_{\text{long}}^{(t)} &= a_{\text{long}}^{(t-1)}
  + C_{\text{throttle}}\,\dot{a}_{\text{long}}\,\Delta t, \label{eq:veh_along} \\
a_{\text{lat}}^{(t)} &= a_{\text{lat}}^{(t-1)}
  + C_{\text{steer}}\,\dot{a}_{\text{lat}}\,\Delta t. \label{eq:veh_alat}
\end{align}
When longitudinal acceleration changes sign
($a_{\text{long}}^{(t-1)} \cdot a_{\text{long}}^{(t)} < 0$), it is snapped
to zero, enabling smooth waiting and constant-velocity trajectories.
A per-episode coefficient $C_{\text{acc}}$ scales the upper acceleration
bound before clamping:
\begin{equation}
a_{\text{long}}^{(t)} \leftarrow
  \mathrm{clip}\!\left(a_{\text{long}}^{(t)},\;
  -5.0,\; 2.5\,C_{\text{acc}}\right), \qquad
a_{\text{lat}}^{(t)} \leftarrow
  \mathrm{clip}\!\left(a_{\text{lat}}^{(t)},\; -4.0,\; 4.0\right).
\label{eq:veh_aclip}
\end{equation}

\paragraph{Velocity Update.}
Speed integrates by the trapezoidal rule and is clipped with a
$C_{\text{vel}}$-scaled upper bound:
\begin{equation}
v^{(t)} = v^{(t-1)} + \tfrac{1}{2}\!\left(a_{\text{long}}^{(t)}
  + a_{\text{long}}^{(t-1)}\right)\Delta t, \qquad
v^{(t)} \leftarrow \mathrm{clip}\!\left(v^{(t)},\; -2.0,\;
  20.0\,C_{\text{vel}}\right).
\label{eq:veh_vel}
\end{equation}
Velocity is snapped to zero on sign change, as with acceleration.

\paragraph{Steering Angle.}
The target steering angle is derived from the desired curvature
$\kappa = a_{\text{lat}}^{(t)}\,/\max\!\left((v^{(t)})^2,\,\epsilon\right)$
with $\epsilon=10^{-5}$:
\begin{align}
\phi_{\text{target}} &= \arctan\!\left(\kappa\, l_{\text{wb}}\right), \label{eq:veh_steer_target} \\
\Delta\phi &= \mathrm{clip}\!\left(\phi_{\text{target}} - \phi^{(t-1)},\;
  -\dot\delta_{\max}\,\Delta t,\; \dot\delta_{\max}\,\Delta t\right), \label{eq:veh_steer_slew} \\
\phi^{(t)} &= \mathrm{clip}\!\left(\phi^{(t-1)}+\Delta\phi,\;
  -\phi_{\max},\;\phi_{\max}\right), \label{eq:veh_steer_clip}
\end{align}
where $l_{\text{wb}} = 0.6 \times \textit{length}$ is the wheelbase,
$\dot\delta_{\max}=0.6$ rad/s is the slew rate limit, and
$\phi_{\max}=0.55$ rad is the hard steering cap.
After clamping, curvature and lateral acceleration are recomputed from the
realized steering angle to maintain consistency:
$\kappa \leftarrow \tan\phi^{(t)}\!/\,l_{\text{wb}}$,
$a_{\text{lat}}^{(t)} \leftarrow (v^{(t)})^2\,\kappa$.

\paragraph{Position Update.}
The pose advances along a circular arc:
\begin{equation}
d = \tfrac{1}{2}(v^{(t)}+v^{(t-1)})\,\Delta t,\qquad
\vartheta = d\,\kappa,\qquad
\Delta x_{\text{local}} = \frac{\sin\vartheta}{\kappa},\qquad
\Delta y_{\text{local}} = \frac{1-\cos\vartheta}{\kappa},
\label{eq:veh_pos}
\end{equation}
rotated into the global frame by the current heading; heading updates as
$\theta^{(t)} = \theta^{(t-1)} + \vartheta$.
When $|\kappa|<10^{-5}$ (nearly straight), the arc degenerates to a
line segment $(\Delta x_{\text{local}},\,\Delta y_{\text{local}})=(d,\,0)$.

\subsection{Pedestrian: Unicycle}

Pedestrian motion is modeled as a unicycle where heading is an
\emph{integrated state} and velocity is constrained to lie along the heading
direction (zero slip).
The action is a pair $(a_v,\,\omega)$, longitudinal acceleration and
heading rate, selected from a nonuniform $7\times7$ grid plus a dedicated
\textsc{stop} action:
\begin{gather*}
a_v \in \{-2.0,\,-0.8,\,-0.2,\,0,\,0.2,\,0.8,\,2.0\}\;\text{m/s}^2,\\
\omega \in \{-2.0,\,-0.5,\,-0.1,\,0,\,0.1,\,0.5,\,2.0\}\;\text{rad/s},
\end{gather*}
giving $|\mathcal{A}|=50$ (49 grid + 1 stop).
The grid concentrates resolution near zero where typical pedestrian
accelerations and heading changes lie (Waymo WOMD $p_{90}$:
$|a_v|\approx1.5$ m/s$^2$, $|\omega|\approx1.3$ rad/s).
The \textsc{stop} action sets both $a_v$ and $\omega$ to zero and forces
speed to zero, providing a clean discrete ``stand still'' primitive.

\paragraph{Integration.}
After clipping $(a_v \leftarrow \mathrm{clip}(a_v,\,-2.0,\,2.0)$,
$\omega \leftarrow \mathrm{clip}(\omega,\,-2.0,\,2.0))$:
\begin{align}
v^{(t)} &= \mathrm{clip}\!\left(v^{(t-1)} + a_v\,\Delta t,\;0,\;3.0\right),
  \label{eq:ped_vel} \\
\theta^{(t)} &= \theta^{(t-1)} + \omega\,\Delta t,
  \label{eq:ped_heading} \\
\mathbf{v}^{(t)} &= v^{(t)}\!\left(\cos\theta^{(t)},\,
  \sin\theta^{(t)}\right),
  \label{eq:ped_vvec} \\
\mathbf{p}^{(t)} &= \mathbf{p}^{(t-1)} + \mathbf{v}^{(t)}\,\Delta t.
  \label{eq:ped_pos}
\end{align}
Speed is clipped to $[0,\,3.0]$ m/s (walking/jogging cap).
Body-frame accelerations for observation and reward parity with other
models are derived numerically:
$a_{\text{long}}=(v^{(t)}-v^{(t-1)})/\Delta t$,
$a_{\text{lat}}=v^{(t)}\cdot(\theta^{(t)}-\theta^{(t-1)})/\Delta t$.

\subsection{Cyclist: Kinematic Bicycle}

Cyclists use a kinematic bicycle model where the action directly controls
acceleration $a$ and front-wheel steering angle $\delta$.
The discrete grid is $7\times7$ with \textbf{logarithmic spacing}:
\begin{gather*}
a \in \{-5.0,\,-2.0,\,-0.5,\,0,\,0.5,\,2.0,\,5.0\}\;\text{m/s}^2,\\
\delta \in \{-0.20,\,-0.06,\,-0.02,\,0,\,0.02,\,0.06,\,0.20\}\;\text{rad},
\end{gather*}
giving $|\mathcal{A}|=49$.
The geometric ratio (${\approx}\,3$) concentrates roughly half of all
actions within the low-magnitude regime where fine steering matters most
for typical urban cycling, providing about $3\times$ finer resolution near
zero for both acceleration and steering than a uniform 7-point grid with the
same bounds.

\paragraph{Scaling and Clipping.}
Per-episode coefficients $C_{\text{steer}}, C_{\text{acc}}, C_{\text{vel}}$
scale the action before integration:
\begin{equation}
\delta \leftarrow C_{\text{steer}}\,\delta,\qquad
a \leftarrow \mathrm{clip}\!\left(\tilde{a},\;-5.0,\;3.0\right),\qquad
\tilde{a} = \begin{cases} C_{\text{acc}}\,a & a > 0,\\ a & a \le 0, \end{cases}
\label{eq:cyc_clip}
\end{equation}
where $C_{\text{acc}}$ scales only forward acceleration (braking is not power-limited).

\paragraph{Velocity Update.}
\begin{equation}
v^{(t)} = v^{(t-1)} + a\,\Delta t,\qquad
v^{(t)} \leftarrow \mathrm{clip}\!\left(v^{(t)},\;
  -1.0,\;10.0\,C_{\text{vel}}\right).
\label{eq:cyc_vel}
\end{equation}
Velocity is snapped to zero on sign change.

\paragraph{Slip Angle and Yaw Rate.}
The slip angle $\beta$ is computed via a $\tanh$-smoothed approximation
that prevents numerical instability at large steering angles:
\begin{equation}
\beta = \tanh\!\left(\tfrac{1}{2}\tan\delta\right),\qquad
\dot\theta = \frac{v^{(t)}\cos\beta\,\tan\delta}{L},
\label{eq:cyc_yaw}
\end{equation}
where $L = \max(\textit{length},\,0.1)$\,m is the effective wheelbase,
guarded against degenerate zero-length entities.

\paragraph{Position and Heading Update.}
\begin{align}
\theta^{(t)} &= \theta^{(t-1)} + \dot\theta\,\Delta t,
  \label{eq:cyc_heading} \\
\mathbf{v}^{(t)} &= v^{(t)}\!\left(
  \cos(\theta^{(t-1)}+\beta),\;\sin(\theta^{(t-1)}+\beta)\right),
  \label{eq:cyc_vvec} \\
\mathbf{p}^{(t)} &= \mathbf{p}^{(t-1)} + \mathbf{v}^{(t)}\,\Delta t.
  \label{eq:cyc_pos}
\end{align}
Note that the velocity direction $\theta^{(t-1)}+\beta$ differs from the
heading $\theta^{(t)}$ by the slip angle during turns.
Lateral acceleration is derived as
$a_{\text{lat}}=v^{(t)}\!\cdot\!\dot\theta$.

\subsection{Domain Randomization Parameters}

Each dynamics model is parameterized by per-episode coefficients
$C_{\text{throttle}}, C_{\text{steer}}, C_{\text{acc}}, C_{\text{vel}}$
sampled uniformly from type-specific ranges at each episode reset
(Table~\ref{tab:kinematic_params}).
Vehicle coefficients span a wide $[0.5,\,1.5]$ band to encourage robustness
across car-to-truck-scale dynamics; cyclist coefficients use a narrower
$[0.7,\,1.3]$ band reflecting the tighter physical envelope of bicycle
kinematics; pedestrian coefficients are fixed at $1.0$ since the unicycle
model already covers the behaviorally relevant range.
All four coefficients are included as observation features so that the
policy can condition on its current kinematic regime.

\begin{table}[!htbp]
\centering
\caption{Per-type kinematic parameters and domain-randomization ranges. Coefficient ranges ($C_*$) are sampled uniformly per agent per episode; scalar values are fixed. Clip bounds are hard limits enforced by the dynamics integrator. Bounding-box ranges are sampled uniformly per agent at spawn, calibrated to the Waymo Open Motion Dataset distribution. Vehicles additionally draw a Bernoulli truck/car bodywork class that selects from an extended length range $[6.1,\,12.2]$\,m for trucks.}
\label{tab:kinematic_params}
\footnotesize
\setlength{\tabcolsep}{4pt}
\begin{tabular}{@{}lccc@{}}
\toprule
Parameter & Vehicle & Pedestrian & Cyclist \\
\midrule
Dynamics model & jerk-Ackermann & unicycle & kin.\ bicycle \\
\midrule
$C_{\text{throttle}}$ & $[0.5,\,1.5]$ & $1.0$ & $1.0$ \\
$C_{\text{steer}}$ & $[0.5,\,1.5]$ & $1.0$ & $[0.7,\,1.3]$ \\
$C_{\text{acc}}$ & $[0.5,\,1.5]$ & $1.0$ & $[0.7,\,1.3]$ \\
$C_{\text{vel}}$ & $[0.5,\,1.5]$ & $1.0$ & $[0.7,\,1.3]$ \\
\midrule
$v$ clip (m/s) & $[-2.0,\,20.0]$ & $[0.0,\,3.0]$ & $[-1.0,\,10.0]$ \\
$a_{\text{long}}$ clip (m/s$^2$) & $[-5.0,\,2.5]$ & $[-2.0,\,2.0]$ & $[-5.0,\,3.0]$ \\
$a_{\text{lat}}$ clip (m/s$^2$) & $[-4.0,\,4.0]$ & $[-4.0,\,4.0]$ & $[-4.0,\,4.0]$ \\
$|\delta|$ cap (rad) & $0.55$ & --- & --- \\
$|\dot\delta|$ slew (rad/s) & $0.6$ & --- & --- \\
$|\omega|$ cap (rad/s) & --- & $2.0$ & --- \\
\midrule
Length (m) & $[3.5,\,5.5]$ & $[0.65,\,1.0]$ & $[1.4,\,2.0]$ \\
Width (m) & $[1.5,\,2.5]$ & $[0.65,\,1.0]$ & $[0.71,\,1.0]$ \\
Height (m) & $[1.2,\,2.0]$ & $[1.5,\,1.9]$ & $[1.5,\,1.9]$ \\
\bottomrule
\end{tabular}
\end{table}

\section{Random Initialization Details}
\label{app:init}

Map memory is partitioned into a \textbf{two-band entity layout}: index range
$[0,N_{\text{log}})$ for log-replay entities and
$[N_{\text{log}},N_{\text{log}}+N_{\text{rand}})$ for randomly placed
controlled agents.
This layout is a simulator capability. The reported SHRED training run
uses only the randomly placed controlled band and does not replay logged future
trajectories.

For each type $k\in\mathcal{K}$ we draw up to $N_k$ agents by
\textbf{rejection sampling}: vehicles snap to lane segments with heading noise
$\Delta\theta\sim\mathcal{U}(-\pi/7.2,\pi/7.2)$, initial speed
$v_0\sim\mathcal{U}(0,2)$ m/s, and a Bernoulli truck/car bodywork draw;
pedestrians sample a crosswalk with probability $p_{\text{cw}}$ and otherwise
a road edge; cyclists use the same lane-sampling pipeline as vehicles.
Each candidate is rejected on collision or off-road, with at most $100$
attempts per slot.

\paragraph{Per-Type Goal Generation.}
When an agent is randomly spawned (or reaches its current goal), a new goal
is synthesized from map topology rather than copied from the log.
The generation strategy is type-specific:
\begin{itemize}
    \item \textbf{Vehicles} draw a target distance
      $d\sim\mathcal{U}(d_{\min},d_{\max})$ and walk forward along the lane
      topology graph, randomly selecting successors at lane ends (up to 10
      transitions), producing goals that follow the road structure.
      Candidate goals are filtered for heading alignment
      (${\geq}\cos 45^{\circ}$).
    \item \textbf{Pedestrians} spawn on either a crosswalk (goal at the
      opposite edge) or a road edge (goal along the edge polyline with a
      sidewalk-side lateral offset), with distance bounds
      $d\in[10,30]$\,m.
      Subsequent goals chain along the nearest road edge using the
      pedestrian's current heading for directionality.
    \item \textbf{Cyclists} follow the same lane-topology walk as vehicles,
      with distance bounds $d\in[20,60]$\,m.
\end{itemize}
To prevent the policy from learning to exploit goal-failure removal as a
shortcut, agents that fail to obtain a valid goal are \emph{not} removed
from the scene.
Instead, they receive an additional binary observation feature indicating
the absence of an active goal, and the policy must learn to act sensibly
under partial supervision.

\section{Per-Type Reward Details}
\label{app:rewards}

Each agent's reward is a weighted sum of terms spanning safety, progress,
comfort, and type-specific constraints.
Let $r^k_t = \sum_i \alpha^k_i\, r_{i,t}$ denote the reward for an agent
of type $k$ at step $t$, where each weight $\alpha^k_i$ is sampled
independently from a type-specific uniform range at episode reset
(reward domain randomization).
Table~\ref{tab:reward_terms} lists the full per-type reward structure.

\paragraph{Comfort.}
The comfort term counts the number of kinematic channels that exceed a
type-specific threshold:
\[
N_{\text{comfort}}=\sum_{i\in\{a_l,\,a_t,\,j_l,\,j_t\}}
  \mathbb{1}\!\left[|x_i| > \tau^{(k)}_i\right],
\]
with $r_{\text{comfort}} = -\alpha_{\text{comfort}}\,N_{\text{comfort}}$.
Table~\ref{tab:comfort} lists the thresholds, calibrated per type from an
80k-scenario analysis of the Waymo Open Motion Dataset at the
$p_{95}$--$p_{97.5}$ percentile of real-data magnitudes.

\paragraph{Pedestrian Road Intrusion.}
Pedestrians receive a per-step jaywalking penalty when their center is on
the road surface and \emph{not} within a crosswalk polygon:
\[
r_{\text{intrusion}} =
  -\alpha_{\text{intrusion}} \cdot \Delta t \cdot
  \mathbb{1}[\text{on\_road}] \cdot \mathbb{1}[\neg\,\text{on\_crosswalk}].
\]
This term is zero for vehicles and cyclists, who are expected to occupy
road lanes.

\paragraph{Type-Specific Lane and Speed Terms.}
Vehicles receive lane alignment ($\alpha_{l\text{-align}}$), lane centering
($\alpha_{l\text{-center}}$), velocity alignment ($\alpha_{v\text{-align}}$),
and reverse penalties ($\alpha_{\text{reverse}}$) that are absent for
pedestrians and cyclists.
Cyclists receive a softer version of lane alignment plus an edge-preference
bonus ($\alpha_{\text{edge}}$) that encourages riding near the road margin.
Both pedestrians and cyclists receive a speed-limit penalty
$r_{\text{speed}} = -\alpha_{\text{speed-lim}} \cdot \Delta t \cdot
(v - v_{\text{lim}})/v_{\max}$ when $v > v_{\text{lim}}$, with
$v_{\text{lim}}$ sampled per agent from $[1.5,\,2.5]$\,m/s for pedestrians and
$[5.0,\,8.0]$\,m/s for cyclists (Table~\ref{tab:reward_terms}).
Similar reward design strategies have been explored in off-road~\citep{wu2026tadpo}
and urban autonomous driving settings~\citep{jaeger2025carl}.

\begin{table}[!htbp]
\centering
\caption{Per-type reward terms and domain-randomization ranges. Dashes mark terms that are disabled ($\alpha=0$) for that type. \emph{Const.}\ values are not randomized; ranges denote uniform sampling per agent per episode. All weights are multiplied by $\Delta t = 0.1$\,s per step except collision and goal, which fire as one-shot events.}
\label{tab:reward_terms}
\footnotesize
\setlength{\tabcolsep}{4pt}
\begin{tabular}{@{}lccc@{}}
\toprule
Reward term & Vehicle & Pedestrian & Cyclist \\
\midrule
$\alpha_{\text{collision}}$ & $[2.0,\,4.0]$ & $[0.5,\,2.0]$ & $[1.0,\,3.0]$ \\
$\alpha_{\text{coll-speed}}$ & $0.1$ & $0.0$ & $0.0$ \\
$m_{k'}$ (partner mult.) & $[1,2,1]$ & --- & --- \\
$\alpha_{\text{boundary}}$ & $[2.0,\,4.0]$ & --- & $[0.5,\,2.0]$ \\
$\alpha_{\text{comfort}}$ & $[0.0,\,0.1]$ & --- & --- \\
$\alpha_{l\text{-align}}$ & $[2.5\text{e-}4,\,2.5\text{e-}2]$ & --- & $[5\text{e-}3,\,2\text{e-}2]$ \\
$\alpha_{v\text{-align}}$ & $[0.0,\,1.0]$ & --- & $[0.25,\,0.75]$ \\
$\alpha_{l\text{-center}}$ & $[2.5\text{e-}4,\,7.5\text{e-}3]$ & --- & --- \\
$\alpha_{\text{reverse}}$ & $[2.5\text{e-}4,\,7.5\text{e-}3]$ & --- & --- \\
$\alpha_{\text{intrusion}}$ & --- & $[0.5,\,2.0]$ & --- \\
$\alpha_{\text{speed-lim}}$ & --- & $[0.5,\,2.0]$ & $[0.25,\,1.0]$ \\
$v_{\text{lim}}$ (m/s) & --- & $[1.5,\,2.5]$ & $[5.0,\,8.0]$ \\
$\alpha_{\text{edge}}$ & --- & --- & $[0.01,\,0.05]$ \\
$\delta_{\text{goal}}$ (m) & $[2.0,\,12.0]$ & $[2.0,\,12.0]$ & $[2.0,\,12.0]$ \\
$\alpha_{\text{goal}}$ & $1.0$ & $1.0$ & $1.0$ \\
\bottomrule
\end{tabular}
\end{table}

\begin{table}[!htbp]
\centering
\caption{Per-type comfort thresholds $\tau^{(k)}$, calibrated at the $p_{95}$--$p_{97.5}$ percentile of the Waymo Open Motion Dataset. Pedestrian and cyclist thresholds are calibrated but unused in the reported run, where $\alpha_{\text{comfort}}=0$ for those types.}
\label{tab:comfort}
\footnotesize
\setlength{\tabcolsep}{4pt}
\begin{tabular}{@{}l cccc@{}}
\toprule
Type & $\tau_{a_l}$ (m/s$^2$) & $\tau_{a_t}$ (m/s$^2$) & $\tau_{j_l}$ (m/s$^3$) & $\tau_{j_t}$ (m/s$^3$) \\
\midrule
Vehicle    & 3.0 & 3.0 & 5.0  & 5.0  \\
Pedestrian & 1.5 & 1.0 & 12.0 & 5.0  \\
Cyclist    & 3.0 & 1.5 & 40.0 & 18.0 \\
\bottomrule
\end{tabular}
\end{table}

\section{PopArt Weight Correction}
\label{app:popart}

Each per-type PopArt layer maintains running statistics
$(\mu_k, \sigma_k, \nu_k)$ updated by EMA over each epoch's batch returns:
\[
\mu_k \leftarrow (1-\beta_k)\mu_k + \beta_k\,\bar{G}_k,\qquad
\nu_k \leftarrow (1-\beta_k)\nu_k + \beta_k\,\overline{G^2}_k,\qquad
\sigma_k \leftarrow \sqrt{\nu_k - \mu_k^{\,2}},
\]
where $\beta_k = 1 - \rho^{T_{\text{bptt}}}$ with configurable decay
$\rho=0.9997$.
To keep the real-scale value function unchanged when $(\mu_k,\sigma_k)$
shifts, the linear weights of $\mathrm{Value}^k$ are corrected:
\[
W^k \leftarrow \frac{\sigma_k^{\text{old}}}{\sigma_k^{\text{new}}}\,W^k,\qquad
b^k \leftarrow
  \frac{\sigma_k^{\text{old}} b^k + \mu_k^{\text{old}} - \mu_k^{\text{new}}}
       {\sigma_k^{\text{new}}},
\]
ensuring no policy-gradient discontinuity at the commit step.

\section{WOSAC Evaluation Protocol and Additional Results}
\label{app:wosac}

\subsection{WOSAC 2024 Validation}

Training remains procedural and does not use logged-trajectory replay.

We compute the realism meta-score with the official weighted definition and prepend logged history when computing kinematic features. On 50 WOSAC scenarios, the GPU evaluator and the official evaluator agree within $3.5\times10^{-5}$ on the realism meta-score.

\begin{table}[t]
\centering
\caption{Agreement between the GPU evaluator and the official WOSAC evaluator on 50 scenarios.}
\label{tab:parity}
\small
\begin{tabular}{@{}lccc@{}}
\toprule
Metric & GPU evaluator & Official evaluator & Maximum difference \\
\midrule
Distance to nearest object & 0.258244 & 0.258244 & $1.2\times10^{-7}$ \\
Time to collision & 0.882676 & 0.882676 & $1.3\times10^{-7}$ \\
Collision indication & 0.850935 & 0.850935 & $8.0\times10^{-8}$ \\
Distance to road edge & 0.646314 & 0.646314 & $1.4\times10^{-7}$ \\
Off-road indication & 0.867736 & 0.867736 & $8.7\times10^{-8}$ \\
Linear speed & 0.236808 & 0.236271 & $3.7\times10^{-3}$ \\
Linear acceleration & 0.365495 & 0.367187 & $1.8\times10^{-2}$ \\
Angular speed & 0.366593 & 0.365808 & $4.0\times10^{-3}$ \\
Angular acceleration & 0.586135 & 0.585067 & $8.1\times10^{-3}$ \\
Realism meta-score & 0.686143 & 0.686108 & $3.5\times10^{-5}$ \\
\bottomrule
\end{tabular}
\end{table}

\subsection{WOSAC 2025 Test Evaluation}

Table~\ref{tab:wosac2025} reports the WOSAC 2025 test evaluation. The 2025 metric differs from the 2023 and 2024 metrics, so we report it separately.

\begin{table}[t]
\centering
\caption{WOSAC 2025 test evaluation for \method.}
\label{tab:wosac2025}
\small
\begin{tabular}{@{}lc@{}}
\toprule
Metric & Value \\
\midrule
Realism meta-score & 0.722 \\
Linear speed & 0.162 \\
Linear acceleration & 0.337 \\
Angular speed & 0.343 \\
Angular acceleration & 0.574 \\
Distance to nearest object & 0.316 \\
Collision & 0.951 \\
Time to collision & 0.806 \\
Distance to road edge & 0.529 \\
Off-road & 0.907 \\
minADE (m) & 5.920 \\
\bottomrule
\end{tabular}
\end{table}

\subsection{Full InterPlan Comparison}
\label{app:interplan_full}

Table~\ref{tab:interplan_full} expands the representative comparison in Table~\ref{tab:interplan} to include all baselines and their listed variants.

\input{tab/tab_interplan_full}

%% file: tab/tab_interplan_full.tex
\begin{table}[!htbp]
\centering
\caption{Full comparison of interactive planners on InterPlan. InterPlan (80) is the official 80-scenario split, and Full Scale (335) evaluates all 335 scenarios. \method achieves the highest score among the compared planners on both evaluation sets. Vehicle-only self-play learns vehicle control with scripted pedestrians and no cyclists.}
\label{tab:interplan_full}
\scriptsize
\setlength{\tabcolsep}{2pt}
\resizebox{\linewidth}{!}{%
\begin{tabular}{@{}llcc@{}}
\toprule
Planner & Type & InterPlan (80) $\uparrow$ & Full Scale (335) $\uparrow$ \\
\midrule
IDM~\citep{treiber2000idm} & Rule & 31 & --- \\
IDM+MOBIL~\citep{treiber2000idm,kesting2007mobil} & Rule & 31 & --- \\
PDM-Closed~\citep{dauner2023pdm} & Rule & 42 & --- \\
SPDM ($N_p{=}15$)~\citep{distelzweig2026spdm} & Rule & 42.00 & --- \\
SPDM ($N_p{=}60$)~\citep{distelzweig2026spdm} & Rule & 63.66 & --- \\
DTPP~\citep{huang2024dtpp} & Hybr. & 25 & --- \\
HybridLLMPlanner (GPT-3.5)~\citep{hallgarten2024interplan} & Hybr. & 40 & --- \\
HybridLLMPlanner (Llama-13B)~\citep{hallgarten2024interplan} & Hybr. & 48 & --- \\
HybridLLMPlanner (Llama-7B)~\citep{hallgarten2024interplan} & Hybr. & 53 & --- \\
PDM+PPO~\citep{behaviorbenchmark2026} & Hybr. & 43.2 & --- \\
FlowDrive w/ guidance + PDM scoring~\citep{wang2025flowdrive} & Hybr. & 44.05 & --- \\
Urban Driver~\citep{scheel2022urbandriver} & IL & 4 & --- \\
GC-PGP~\citep{hallgarten2023gcpgp} & IL & 10 & --- \\
GameFormer~\citep{huang2023gameformer} & IL & 11 & --- \\
PDM-Open~\citep{dauner2023pdm} & IL & 25 & --- \\
Diffusion Planner~\citep{zheng2025diffplanner} & IL & 25.8 & --- \\
FlowDrive~\citep{wang2025flowdrive} & IL & 36.96 & --- \\
Flow Planner~\citep{tan2025flowplanner} & IL & --- & 61.82 \\
G2DP~\citep{yu2026g2dp} & IL & --- & 61.74 \\
PPO~\citep{behaviorbenchmark2026} & RL & 42.1 & --- \\
\midrule
\method & RL & \textbf{67.50} & \textbf{70.14} \\
\method (vehicle-only) & RL & 54.19 & 66.65 \\
\bottomrule
\end{tabular}%
}
\end{table}

%% file: references.bib
@inproceedings{montali2023wosac,
  title={The {Waymo} Open Sim Agents Challenge},
  author={Montali, Nico and Lambert, John and Mougin, Paul and Kuefler, Alex and Rhinehart, Nicholas and Li, Michelle and Gulino, Cole and Emrich, Tristan and Yang, Zoey and Whiteson, Shimon and White, Brandyn and Anguelov, Dragomir},
  booktitle={Advances in Neural Information Processing Systems (NeurIPS), Datasets and Benchmarks Track},
  volume={36},
  pages={59151--59171},
  year={2023}
}

@inproceedings{cusumano2025gigaflow,
  title={Robust Autonomy Emerges from Self-Play},
  author={Cusumano-Towner, Marco and Hafner, David and Hertzberg, Alexander and Huval, Brody and Petrenko, Aleksei and Vinitsky, Eugene and Wijmans, Erik and Killian, Taylor W. and Bowers, Stuart and Sener, Ozan and Kraehenbuehl, Philipp and Koltun, Vladlen},
  booktitle={Proceedings of the 42nd International Conference on Machine Learning},
  series={Proceedings of Machine Learning Research},
  volume={267},
  pages={11710--11737},
  year={2025},
  publisher={PMLR}
}

@inproceedings{chang2025spacer,
  title={{SPACeR}: Self-Play Anchoring with Centralized Reference Models},
  author={Chang, Wei-Jer and Rangesh, Akshay and Joseph, Kevin and Strong, Matthew and Tomizuka, Masayoshi and Hu, Yihan and Zhan, Wei},
  booktitle={International Conference on Learning Representations (ICLR)},
  year={2026}
}

@inproceedings{wu2024smart,
  title={{SMART}: Scalable Multi-Agent Real-Time Motion Generation via Next-Token Prediction},
  author={Wu, Wei and Feng, Xiaoxin and Gao, Ziyan and Kan, Yuheng},
  booktitle={Advances in Neural Information Processing Systems},
  volume={37},
  pages={114048--114071},
  year={2024}
}

@inproceedings{zhang2025catk,
  title={Closed-Loop Supervised Fine-Tuning of Tokenized Traffic Models},
  author={Zhang, Zhejun and Karkus, Peter and Igl, Maximilian and Ding, Wenhao and Chen, Yuxiao and Ivanovic, Boris and Pavone, Marco},
  booktitle={Proceedings of the IEEE/CVF Conference on Computer Vision and Pattern Recognition},
  pages={5422--5432},
  year={2025}
}

@article{cornelisse2025reliable,
  title={Building Reliable Sim Driving Agents by Scaling Self-Play},
  author={Cornelisse, Daphne and Pandya, Aarav and Joseph, Kevin and Su{\'a}rez, Joseph and Vinitsky, Eugene},
  journal={arXiv preprint arXiv:2502.14706},
  year={2025}
}

@article{cornelisse2024hrppo,
  title={Human-Compatible Driving Partners through Data-Regularized Self-Play Reinforcement Learning},
  author={Cornelisse, Daphne and Vinitsky, Eugene},
  journal={Reinforcement Learning Journal},
  volume={5},
  pages={2320--2344},
  year={2024}
}

@inproceedings{kazemkhani2024gpudrive,
  title={{GPUDrive}: Data-Driven, Multi-Agent Driving Simulation at 1 Million FPS},
  author={Kazemkhani, Saman and Pandya, Aarav and Cornelisse, Daphne and Shacklett, Brennan and Vinitsky, Eugene},
  booktitle={International Conference on Learning Representations (ICLR)},
  year={2025}
}

@inproceedings{jiang2024scenediffuser,
  title={{SceneDiffuser}: Efficient and Controllable Driving Simulation Initialization and Rollout},
  author={Jiang, Max and Bai, Yijing and Cornman, Andre and Davis, Christopher and Huang, Xiukun and Jeon, Hong and Kulshrestha, Sakshum and Lambert, John and Li, Shuangyu and Zhou, Xuanyu and others},
  booktitle={Advances in Neural Information Processing Systems},
  volume={37},
  pages={55729--55760},
  year={2024}
}

@inproceedings{xu2023bits,
  title={{BITS}: Bi-Level Imitation for Traffic Simulation},
  author={Xu, Danfei and Chen, Yuxiao and Ivanovic, Boris and Pavone, Marco},
  booktitle={IEEE International Conference on Robotics and Automation (ICRA)},
  year={2023}
}

@inproceedings{peng2024improving,
  title={Improving Agent Behaviors with {RL} Fine-Tuning for Autonomous Driving},
  author={Peng, Zhenghao and Luo, Wenjie and Lu, Yiren and Shen, Tianyi and Gulino, Cole and Seff, Ari and Fu, Justin},
  booktitle={European Conference on Computer Vision},
  pages={165--181},
  year={2024},
  publisher={Springer}
}

@article{chen2025rift,
  title={{RIFT}: Group-Relative {RL} Fine-Tuning for Realistic and Controllable Traffic Simulation},
  author={Chen, Keyu and Sun, Wenchao and Cheng, Hao and Zheng, Sifa},
  journal={arXiv preprint arXiv:2505.03344},
  year={2025}
}

@inproceedings{cao2024rlhf,
  title={Reinforcement Learning with Human Feedback for Realistic Traffic Simulation},
  author={Cao, Yulong and Ivanovic, Boris and Xiao, Chaowei and Pavone, Marco},
  booktitle={IEEE International Conference on Robotics and Automation (ICRA)},
  pages={14428--14434},
  year={2024}
}

@article{silver2017mastering,
  title={Mastering the Game of {Go} without Human Knowledge},
  author={Silver, David and Schrittwieser, Julian and Simonyan, Karen and Antonoglou, Ioannis and Huang, Aja and Guez, Arthur and Hubert, Thomas and Baker, Lucas and Lai, Matthew and Bolton, Adrian and others},
  journal={Nature},
  volume={550},
  number={7676},
  pages={354--359},
  year={2017}
}

@article{vinyals2019grandmaster,
  title={Grandmaster Level in {StarCraft II} Using Multi-Agent Reinforcement Learning},
  author={Vinyals, Oriol and Babuschkin, Igor and Czarnecki, Wojciech M and Mathieu, Micha{\"e}l and Dudzik, Andrew and Chung, Junyoung and Choi, David H and Powell, Richard and Ewalds, Timo and Georgiev, Petko and others},
  journal={Nature},
  volume={575},
  number={7782},
  pages={350--354},
  year={2019}
}

@article{berner2019dota,
  title={Dota 2 with Large Scale Deep Reinforcement Learning},
  author={Berner, Christopher and Brockman, Greg and Chan, Brooke and Cheung, Vicki and D{\k{e}}biak, Przemys{\l}aw and Dennison, Christy and Farhi, David and Fischer, Quirin and Hashme, Shariq and Hesse, Chris and others},
  journal={arXiv preprint arXiv:1912.06680},
  year={2019}
}

@article{jaderberg2019human,
  title={Human-Level Performance in 3{D} Multiplayer Games with Population-Based Reinforcement Learning},
  author={Jaderberg, Max and Czarnecki, Wojciech M and Dunning, Iain and Marris, Luke and Lever, Guy and Casta{\~n}eda, Antonio Garcia and Beattie, Charles and Rabinowitz, Neil C and Morcos, Ari S and Ruderman, Avraham and others},
  journal={Science},
  volume={364},
  number={6443},
  pages={859--865},
  year={2019}
}

@inproceedings{wang2020roma,
  title={{ROMA}: Multi-Agent Reinforcement Learning with Emergent Roles},
  author={Wang, Tonghan and Dong, Heng and Lesser, Victor and Zhang, Chongjie},
  booktitle={Proceedings of the 37th International Conference on Machine Learning},
  series={Proceedings of Machine Learning Research},
  volume={119},
  pages={9876--9886},
  year={2020},
  publisher={PMLR}
}

@inproceedings{hallgarten2024interplan,
  title={Can Vehicle Motion Planning Generalize to Realistic Long-tail Scenarios?},
  author={Hallgarten, Marcel and Zapata, Julian and Stoll, Martin and Renz, Katrin and Zell, Andreas},
  booktitle={IEEE/RSJ International Conference on Intelligent Robots and Systems (IROS)},
  year={2024}
}

@inproceedings{distelzweig2026spdm,
  title={Perfect Prediction or Plenty of Proposals? What Matters Most in Planning for Autonomous Driving},
  author={Distelzweig, Aron and Janjo{\v{s}}, Faris and Scheel, Oliver and Varra, Sirish Reddy and Rajan, Raghu and Boedecker, Joschka},
  booktitle={IEEE International Conference on Robotics and Automation (ICRA)},
  year={2026},
  note={arXiv:2510.15505}
}

@inproceedings{philion2024trajeglish,
  title={Trajeglish: Traffic Modeling as Next-Token Prediction},
  author={Philion, Jonah and Peng, Xue Bin and Fidler, Sanja},
  booktitle={International Conference on Learning Representations (ICLR)},
  year={2024}
}

@inproceedings{zhou2024behaviorgpt,
  title={{BehaviorGPT}: Smart Agent Simulation for Autonomous Driving with Next-Patch Prediction},
  author={Zhou, Zikang and Hu, Haibo and Chen, Xinhong and Wang, Jianping and Guan, Nan and Wu, Kui and Li, Yung-Hui and Huang, Yu-Kai and Ho, Chun Jason},
  booktitle={Advances in Neural Information Processing Systems},
  year={2024}
}

@inproceedings{tan2024prosim,
  title={Promptable Closed-loop Traffic Simulation},
  author={Tan, Shuhan and Ivanovic, Boris and Chen, Yuxiao and Li, Boyi and Weng, Xinshuo and Cao, Yulong and Kr{\"a}henb{\"u}hl, Philipp and Pavone, Marco},
  booktitle={Conference on Robot Learning (CoRL)},
  year={2024}
}

@inproceedings{pei2025smartr1,
  title={Advancing Multi-Agent Traffic Simulation via {R1}-Style Reinforcement Fine-Tuning},
  author={Pei, Muleilan and Shi, Shaoshuai and Shen, Shaojie},
  booktitle={International Conference on Learning Representations (ICLR)},
  year={2026}
}

@article{yu2022mappo,
  title={The Surprising Effectiveness of {PPO} in Cooperative, Multi-Agent Games},
  author={Yu, Chao and Velu, Akash and Vinitsky, Eugene and Gao, Jiaxuan and Wang, Yu and Bayen, Alexandre and Wu, Yi},
  journal={Advances in Neural Information Processing Systems, Datasets and Benchmarks Track},
  volume={35},
  year={2022}
}

@inproceedings{rashid2018qmix,
  title={{QMIX}: Monotonic Value Function Factorisation for Deep Multi-Agent Reinforcement Learning},
  author={Rashid, Tabish and Samvelyan, Mikayel and Schroeder de Witt, Christian and Farquhar, Gregory and Foerster, Jakob and Whiteson, Shimon},
  booktitle={International Conference on Machine Learning (ICML)},
  year={2018}
}

@inproceedings{kuba2022happo,
  title={Trust Region Policy Optimisation in Multi-Agent Reinforcement Learning},
  author={Kuba, Jakub Grudzien and Chen, Ruiqing and Wen, Muning and Wen, Ying and Sun, Fanglei and Wang, Jun and Yang, Yaodong},
  booktitle={International Conference on Learning Representations (ICLR)},
  year={2022}
}

@article{zhong2024harl,
  title={Heterogeneous-Agent Reinforcement Learning},
  author={Zhong, Yifan and Kuba, Jakub Grudzien and Feng, Xidong and Hu, Siyi and Ji, Jiaming and Yang, Yaodong},
  journal={Journal of Machine Learning Research},
  volume={25},
  year={2024}
}

@inproceedings{christianos2021seps,
  title={Scaling Multi-Agent Reinforcement Learning with Selective Parameter Sharing},
  author={Christianos, Filippos and Papoudakis, Georgios and Rahman, Arrasy and Albrecht, Stefano V},
  booktitle={Advances in Neural Information Processing Systems},
  year={2021}
}

@inproceedings{wang2021rode,
  title={{RODE}: Learning Roles to Decompose Multi-Agent Tasks},
  author={Wang, Tonghan and Gupta, Tarun and Mahajan, Anuj and Peng, Bei and Whiteson, Shimon and Zhang, Chongjie},
  booktitle={International Conference on Learning Representations (ICLR)},
  year={2021}
}

@inproceedings{yang2020softmod,
  title={Multi-Task Reinforcement Learning with Soft Modularization},
  author={Yang, Ruihan and Xu, Huazhe and Wu, Yi and Wang, Xiaolong},
  booktitle={Advances in Neural Information Processing Systems},
  year={2020}
}

@inproceedings{obandoceron2024moe,
  title={Mixtures of Experts Unlock Parameter Scaling for Deep {RL}},
  author={Obando-Ceron, Johan and Sokar, Ghada and Willi, Timon and Lyle, Clare and Farebrother, Jesse and Foerster, Jakob and Dziugaite, Gintare Karolina and Precup, Doina and Castro, Pablo Samuel},
  booktitle={International Conference on Machine Learning (ICML)},
  year={2024}
}

@inproceedings{vanhasselt2016popart,
  title={Learning Values across Many Orders of Magnitude},
  author={van Hasselt, Hado P and Guez, Arthur and Hessel, Matteo and Mnih, Volodymyr and Silver, David},
  booktitle={Advances in Neural Information Processing Systems},
  year={2016}
}

@article{kupper2023social,
  title={Identification of social groups and waiting pedestrians at railway platforms using trajectory data},
  author={K{\"u}pper, Michael and Seyfried, Armin},
  journal={PLoS ONE},
  volume={18},
  number={3},
  pages={e0282526},
  year={2023},
  doi={10.1371/journal.pone.0282526}
}

@article{luquesanchez2020crowd,
  title={Revisiting Crowd Behaviour Analysis Through Deep Learning: Taxonomy, Anomaly Detection, Crowd Emotions, Datasets, Opportunities and Prospects},
  author={Luque S{\'a}nchez, Francisco and Hupont, Isabelle and Tabik, Siham and Herrera, Francisco},
  journal={Information Fusion},
  volume={64},
  pages={318--335},
  year={2020},
  doi={10.1016/j.inffus.2020.07.008}
}

@article{helbing1995social,
  title={Social Force Model for Pedestrian Dynamics},
  author={Helbing, Dirk and Moln{\'a}r, P{\'e}ter},
  journal={Physical Review E},
  volume={51},
  number={5},
  pages={4282--4286},
  year={1995}
}

@article{moussaid2011simple,
  title={How Simple Rules Determine Pedestrian Behavior and Crowd Disasters},
  author={Moussa{\"\i}d, Mehdi and Helbing, Dirk and Theraulaz, Guy},
  journal={Proceedings of the National Academy of Sciences (PNAS)},
  volume={108},
  number={17},
  pages={6884--6888},
  year={2011}
}

@article{helbing2005crowd,
  title={Self-Organized Pedestrian Crowd Dynamics: Experiments, Simulations, and Design Solutions},
  author={Helbing, Dirk and Buzna, Lubo{\v{s}} and Johansson, Anders and Werner, Torsten},
  journal={Transportation Science},
  volume={39},
  number={1},
  pages={1--24},
  year={2005}
}

@incollection{vandenberg2008orca,
  title={Reciprocal $n$-body Collision Avoidance},
  author={van den Berg, Jur and Guy, Stephen J and Lin, Ming and Manocha, Dinesh},
  booktitle={Robotics Research (Springer Tracts in Advanced Robotics)},
  pages={3--19},
  year={2011},
  publisher={Springer}
}

@inproceedings{alahi2016sociallstm,
  title={Social {LSTM}: Human Trajectory Prediction in Crowded Spaces},
  author={Alahi, Alexandre and Goel, Kratarth and Ramanathan, Vignesh and Robicquet, Alexandre and Fei-Fei, Li and Savarese, Silvio},
  booktitle={Proceedings of the IEEE Conference on Computer Vision and Pattern Recognition (CVPR)},
  pages={961--971},
  year={2016}
}

@inproceedings{gupta2018socialgan,
  title={Social {GAN}: Socially Acceptable Trajectories with Generative Adversarial Networks},
  author={Gupta, Agrim and Johnson, Justin and Fei-Fei, Li and Savarese, Silvio and Alahi, Alexandre},
  booktitle={Proceedings of the IEEE Conference on Computer Vision and Pattern Recognition (CVPR)},
  pages={2255--2264},
  year={2018}
}

@inproceedings{salzmann2020trajectronpp,
  title={{Trajectron++}: Dynamically-Feasible Trajectory Forecasting with Heterogeneous Data},
  author={Salzmann, Tim and Ivanovic, Boris and Chakravarty, Punarjay and Pavone, Marco},
  booktitle={European Conference on Computer Vision (ECCV)},
  pages={683--700},
  year={2020}
}

@inproceedings{xu2022socialvae,
  title={{SocialVAE}: Human Trajectory Prediction Using Timewise Latents},
  author={Xu, Pei and Hayet, Jean-Bernard and Karamouzas, Ioannis},
  booktitle={European Conference on Computer Vision (ECCV)},
  pages={511--528},
  year={2022}
}

@inproceedings{rasouli2023pedformer,
  title={{PedFormer}: Pedestrian Behavior Prediction via Cross-Modal Attention Modulation and Gated Multitask Learning},
  author={Rasouli, Amir and Kotseruba, Iuliia},
  booktitle={IEEE International Conference on Robotics and Automation (ICRA)},
  year={2023}
}

@inproceedings{suo2021trafficsim,
  title={{TrafficSim}: Learning to Simulate Realistic Multi-Agent Behaviors},
  author={Suo, Simon and Regalado, Sebastian and Casas, Sergio and Urtasun, Raquel},
  booktitle={Proceedings of the IEEE/CVF Conference on Computer Vision and Pattern Recognition (CVPR)},
  pages={10400--10409},
  year={2021}
}

@inproceedings{vinitsky2022nocturne,
  title={{Nocturne}: A Scalable Driving Benchmark for Bringing Multi-Agent Learning One Step Closer to the Real World},
  author={Vinitsky, Eugene and Lichtl{\'e}, Nathan and Yang, Xiaomeng and Amos, Brandon and Foerster, Jakob},
  booktitle={Advances in Neural Information Processing Systems, Datasets and Benchmarks Track},
  year={2022}
}

@inproceedings{gulino2023waymax,
  title={{Waymax}: An Accelerated, Data-Driven Simulator for Large-Scale Autonomous Driving Research},
  author={Gulino, Cole and Fu, Justin and Luo, Wenjie and Tucker, George and Bronstein, Eli and Lu, Yiren and Harb, Jean and Pan, Xinlei and Wang, Yan and Chen, Xiangyu and others},
  booktitle={Advances in Neural Information Processing Systems, Datasets and Benchmarks Track},
  year={2023}
}

@article{li2022metadrive,
  title={{MetaDrive}: Composing Diverse Driving Scenarios for Generalizable Reinforcement Learning},
  author={Li, Quanyi and Peng, Zhenghao and Feng, Lan and Zhang, Qihang and Xue, Zhenghai and Zhou, Bolei},
  journal={IEEE Transactions on Pattern Analysis and Machine Intelligence},
  year={2022}
}

@inproceedings{rowe2024ctrlsim,
  title={{CtRL-Sim}: Reactive and Controllable Driving Agents with Offline Reinforcement Learning},
  author={Rowe, Luke and Girgis, Roger and Gosselin, Anthony and Carrez, Bruno and Golemo, Florian and Heide, Felix and Paull, Liam and Pal, Christopher},
  booktitle={Conference on Robot Learning (CoRL)},
  year={2024}
}

@article{wang2026nomad,
  title={Learning to Drive in New Cities Without Human Demonstrations},
  author={Wang, Zilin and Rahmani, Saeed and Cornelisse, Daphne and Sarkar, Bidipta and Goldie, Alexander David and Foerster, Jakob Nicolaus and Whiteson, Shimon},
  journal={arXiv preprint arXiv:2602.15891},
  year={2026}
}

@inproceedings{espeholt2018impala,
  title={{IMPALA}: Scalable Distributed Deep-{RL} with Importance Weighted Actor-Learner Architectures},
  author={Espeholt, Lasse and Soyer, Hubert and Munos, R{\'e}mi and Simonyan, Karen and Mnih, Volodymyr and Ward, Tom and Doron, Yotam and Firoiu, Vlad and Harley, Tim and Dunning, Iain and others},
  booktitle={International Conference on Machine Learning (ICML)},
  pages={1407--1416},
  year={2018}
}

@article{wu2026tadpo,
  title={{TADPO}: Reinforcement Learning Goes Off-road},
  author={Wu, Zhouchonghao and Song, Raymond and Mundheda, Vedant and Navarro-Serment, Luis E. and Schoenborn, Christof and Schneider, Jeff},
  journal={arXiv preprint arXiv:2603.05995},
  year={2026}
}

@article{jaeger2025carl,
  title={{CaRL}: Learning Scalable Planning Policies with Simple Rewards},
  author={Jaeger, Bernhard and Dauner, Daniel and Bei{\ss}wenger, Jens and Gerstenecker, Simon and Chitta, Kashyap and Geiger, Andreas},
  journal={arXiv preprint arXiv:2504.17838},
  year={2025}
}

@inproceedings{ahmadi2026rlftsim,
  title={{RLFTSim}: Realistic and Controllable Multi-Agent Traffic Simulation via Reinforcement Learning Fine-Tuning},
  author={Ahmadi, Ehsan and Schofield, Hunter and Khamidehi, Behzad and Arasteh, Fazel and Shan, Jinjun and Mou, Lili and Bai, Dongfeng and Rezaee, Kasra},
  booktitle={Proceedings of the IEEE/CVF Conference on Computer Vision and Pattern Recognition},
  pages={39734--39743},
  year={2026}
}

@article{guo2026decompgail,
  title={{DecompGAIL}: Learning Realistic Traffic Behaviors with Decomposed Multi-Agent Generative Adversarial Imitation Learning},
  author={Guo, Ke and Liu, Haochen and Wu, Xiaojun and Lv, Chen},
  journal={arXiv preprint arXiv:2510.06913},
  year={2026},
  note={Accepted to ICLR}
}

@article{treiber2000idm,
  title={Congested Traffic States in Empirical Observations and Microscopic Simulations},
  author={Treiber, Martin and Hennecke, Ansgar and Helbing, Dirk},
  journal={Physical Review E},
  volume={62},
  pages={1805--1824},
  year={2000}
}

@article{kesting2007mobil,
  title={General Lane-Changing Model {MOBIL} for Car-Following Models},
  author={Kesting, Arne and Treiber, Martin and Helbing, Dirk},
  journal={Transportation Research Record},
  volume={1999},
  number={1},
  pages={86--94},
  year={2007}
}

@inproceedings{dauner2023pdm,
  title={Parting with Misconceptions about Learning-based Vehicle Motion Planning},
  author={Dauner, Daniel and Hallgarten, Marcel and Geiger, Andreas and Chitta, Kashyap},
  booktitle={Conference on Robot Learning (CoRL)},
  year={2023}
}

@inproceedings{huang2024dtpp,
  title={{DTPP}: Differentiable Joint Conditional Prediction and Cost Evaluation for Tree Policy Planning in Autonomous Driving},
  author={Huang, Zhiyu and Karkus, Peter and Ivanovic, Boris and Chen, Yuxiao and Pavone, Marco and Lv, Chen},
  booktitle={IEEE International Conference on Robotics and Automation (ICRA)},
  year={2024}
}

@article{behaviorbenchmark2026,
  title={Beyond Self-Play and Scale: A Behavior Benchmark for Generalization in Autonomous Driving},
  author={Distelzweig, Aron and Janjo{\v{s}}, Faris and Look, Andreas and Rothenh{\"a}usler, Anna and Jost, Daniel and Scheel, Oliver and Rajan, Raghu and Cornelisse, Daphne and Vinitsky, Eugene and Boedecker, Joschka},
  journal={arXiv preprint arXiv:2605.10034},
  year={2026}
}

@article{wang2025flowdrive,
  title={{FlowDrive}: Moderated Flow Matching with Data Balancing for Trajectory Planning},
  author={Wang, Lingguang and Ta{\c{s}}, {\"O}mer {\c{S}}ahin and Steiner, Marlon and Stiller, Christoph},
  journal={arXiv preprint arXiv:2509.21961},
  year={2025}
}

@inproceedings{scheel2022urbandriver,
  title={Urban Driver: Learning to Drive from Real-world Demonstrations Using Policy Gradients},
  author={Scheel, Oliver and Bergamini, Luca and Wolczyk, Maciej and Osi{\'n}ski, B{\l}a{\.z}ej and Ondruska, Peter},
  booktitle={Conference on Robot Learning (CoRL)},
  pages={718--728},
  year={2022}
}

@inproceedings{hallgarten2023gcpgp,
  title={From Prediction to Planning with Goal Conditioned Lane Graph Traversals},
  author={Hallgarten, Marcel and Stoll, Martin and Zell, Andreas},
  booktitle={IEEE International Conference on Intelligent Transportation Systems (ITSC)},
  year={2023}
}

@inproceedings{huang2023gameformer,
  title={{GameFormer}: Game-theoretic Modeling and Learning of Transformer-based Interactive Prediction and Planning for Autonomous Driving},
  author={Huang, Zhiyu and Liu, Haochen and Lv, Chen},
  booktitle={IEEE/CVF International Conference on Computer Vision (ICCV)},
  year={2023}
}

@inproceedings{zheng2025diffplanner,
  title={Diffusion-Based Planning for Autonomous Driving with Flexible Guidance},
  author={Zheng, Yinan and Liang, Ruiming and Zheng, Kexin and Zheng, Jinliang and Mao, Liyuan and Li, Jianxiong and Gu, Weihao and Ai, Rui and Li, Shengbo Eben and Zhan, Xianyuan and Liu, Jingjing},
  booktitle={International Conference on Learning Representations (ICLR)},
  year={2025},
  note={arXiv:2501.15564}
}

@inproceedings{tan2025flowplanner,
  title={Flow Matching-Based Autonomous Driving Planning with Advanced Interactive Behavior Modeling},
  author={Tan, Tianyi and Zheng, Yinan and Liang, Ruiming and Wang, Zexu and Zheng, Kexin and Zheng, Jinliang and Li, Jianxiong and Zhan, Xianyuan and Liu, Jingjing},
  booktitle={Advances in Neural Information Processing Systems (NeurIPS)},
  year={2025},
  note={arXiv:2510.11083}
}

@article{yu2026g2dp,
  title={{G2DP}: Diffusion Planning with Spatio-Temporal Grid Guidance},
  author={Yu, Hang and Jin, Ye and Canevaro, Alessandro and Schmidt, Julian and Jordan, Julian and Li, Peizheng and Kaufeld, Marc and Lindner, Silvan and Betz, Johannes and Stork, Wilhelm},
  journal={arXiv preprint arXiv:2606.26017},
  year={2026}
}

@misc{wu2026terrazero,
  title = {{TerraZero}: Procedural Driving Simulation for Zero-Demonstration Self-Play at Scale},
  author = {Zhouchonghao Wu and Akshay Rangesh and Weixin Li and Wei-Jer Chang and Zachary Lee and Saeed Bonab and Tim Wang and Wei Zhan},
  year = {2026},
  eprint = {2607.13028},
  archivePrefix = {arXiv},
  primaryClass = {cs.LG},
  url = {https://arxiv.org/abs/2607.13028}
}

@misc{xiong2026terratransfer,
  title = {{TerraTransfer}: Learning End-to-End Driving Policies Without Expert Demonstrations},
  author = {Zikang Xiong and Weixin Li and Zhouchonghao Wu and Akshay Rangesh and Saarth Bonde and Grantland Hall and Chen Tang and Yihan Hu and Wei Zhan},
  year = {2026},
  eprint = {2606.17386},
  archivePrefix = {arXiv},
  primaryClass = {cs.CV},
  url = {https://arxiv.org/abs/2606.17386}
}
